\documentclass[10pt,journal,compsoc]{IEEEtran}

\usepackage{color}
\usepackage{times}
\usepackage{epsfig}
\usepackage{graphicx}
\usepackage{subfigure}
\usepackage{amsmath}
\usepackage{amssymb}

\usepackage{mathrsfs}
\usepackage{algorithm}
\usepackage{algorithmic}
\usepackage{array}
\usepackage{booktabs}
\usepackage{multirow}

\newcommand{\INPUT}{\item[\myinput]}
\newcommand{\myinput}{\textbf{Preparation:}}

\newcommand{\MYWHILE}{\item[\mywhile]}
\newcommand{\mywhile}{\textbf{repeat}}

\newcommand{\MYENDWHILE}{\item[\myendwhile]}
\newcommand{\myendwhile}{\textbf{until}}

\usepackage{ifpdf}

\ifCLASSINFOpdf
  \DeclareGraphicsExtensions{.pdf,.jpeg,.png}
    \DeclareGraphicsExtensions{.eps}
\usepackage{epstopdf}
\else
\fi

\begin{document}
%
\title{Hierarchical Scene Parsing by Weakly Supervised Learning with Image Descriptions}


\author{Ruimao Zhang, Liang Lin, Guangrun Wang, Meng Wang, and Wangmeng Zuo
\IEEEcompsocitemizethanks {\IEEEcompsocthanksitem
R. Zhang, L. Lin and G. Wang are with the School of Data and Computer Science, Sun Yat-sen University, Guangzhou, P. R. China (E-mail: ruimao.zhang@ieee.org; linliang@ieee.org; wanggrun@mail2.sysu.edu.cn). Corresponding author is Liang Lin.
\IEEEcompsocthanksitem
M. Wang is with the School of Computer Science
and Information Engineering, Hefei University of Technology, Hefei, P. R. China (E-mail: eric.mengwang@gmail.com).
\IEEEcompsocthanksitem
W. Zuo is with the School of Computer Science, Harbin Institute of Technology, Harbin, P. R. China (E-mail: cswmzuo@gmail.com).}
}

\markboth{IEEE TRANSACTIONS ON PATTERN ANALYSIS AND MACHINE INTELLIGENCE}%
{Shell \MakeLowercase{\textit{et al.}}: Bare Demo of IEEEtran.cls for IEEE Transactions on Magnetics Journals}

\IEEEcompsoctitleabstractindextext{
\begin{abstract}
This paper investigates a fundamental problem of scene understanding: how to parse a scene image into a structured configuration (i.e., a semantic object hierarchy with object interaction relations). We propose a deep architecture consisting of two networks: i) a convolutional neural network (CNN) extracting the image representation for pixel-wise object labeling and ii) a recursive neural network (RsNN) discovering the hierarchical object structure and the inter-object relations. Rather than relying on elaborative annotations (e.g., manually labeled semantic maps and relations), we train our deep model in a weakly-supervised learning manner by leveraging the descriptive sentences of the training images. Specifically, we decompose each sentence into a semantic tree consisting of nouns and verb phrases, and apply these tree structures to discover the configurations of the training images.
Once these scene configurations are determined, then the parameters of both the CNN and RsNN are updated accordingly by back propagation. The entire model training is accomplished through an Expectation-Maximization method. Extensive experiments show that our model is capable of producing meaningful scene configurations and achieving more favorable scene labeling results on two benchmarks (i.e., PASCAL VOC 2012 and SYSU-Scenes) compared with other state-of-the-art weakly-supervised deep learning methods.
In particular, SYSU-Scenes contains more than 5000 scene images with their semantic sentence descriptions, which is created by us for advancing research on scene parsing.

\end{abstract}

}

\maketitle

\IEEEdisplaynontitleabstractindextext

%
\IEEEpeerreviewmaketitle

\section{Introduction}
Scene understanding started with the goal of creating systems that can infer meaningful configurations (e.g., parts, objects and their compositions with relations) from imagery like humans~\cite{DBLP:ImageParsing-Attribute}\cite{J:Space-Tiling}. In computer vision research, most of the scene understanding methods focus on semantic scene labeling / segmentation problems (e.g., assigning semantic labels to each pixel)~\cite{DBLP:SemanSeg-1}\cite{DBLP:SemanSeg-3}\cite{DBLP:FCnetwork}\cite{DBLP:RecursiveContext}. Yet relatively few works attempt to explore how to automatically generate a structured and meaningful configuration of the input scene, which is an essential task to human cognition~\cite{B:CognitionScience}. In spite of some acknowledged structured models beyond scene labeling, e.g., and-or graph (AoG)~\cite{J:AoG}, factor graph (FG)~\cite{J:FactorGraphs} and recursive neural network (RsNN)~\cite{DBLP:Recursive_Socher}, learning the hierarchical scene structure remains a challenge due to the following difficulties.

\begin{itemize}

\item The parsing configurations of nested hierarchical structure in scene images are often ambiguous, e.g., a configuration may have more than one parse. Moreover, making the parsing result in accordance with human perception is also intractable.

\item Training a scene parsing model usually relies on very expensive manual annotations, e.g., labeling pixel-wise semantic maps, hierarchical representations and inter-object relations.

\end{itemize}

\begin{figure}[t]
\centering
\includegraphics[width=3.2in]{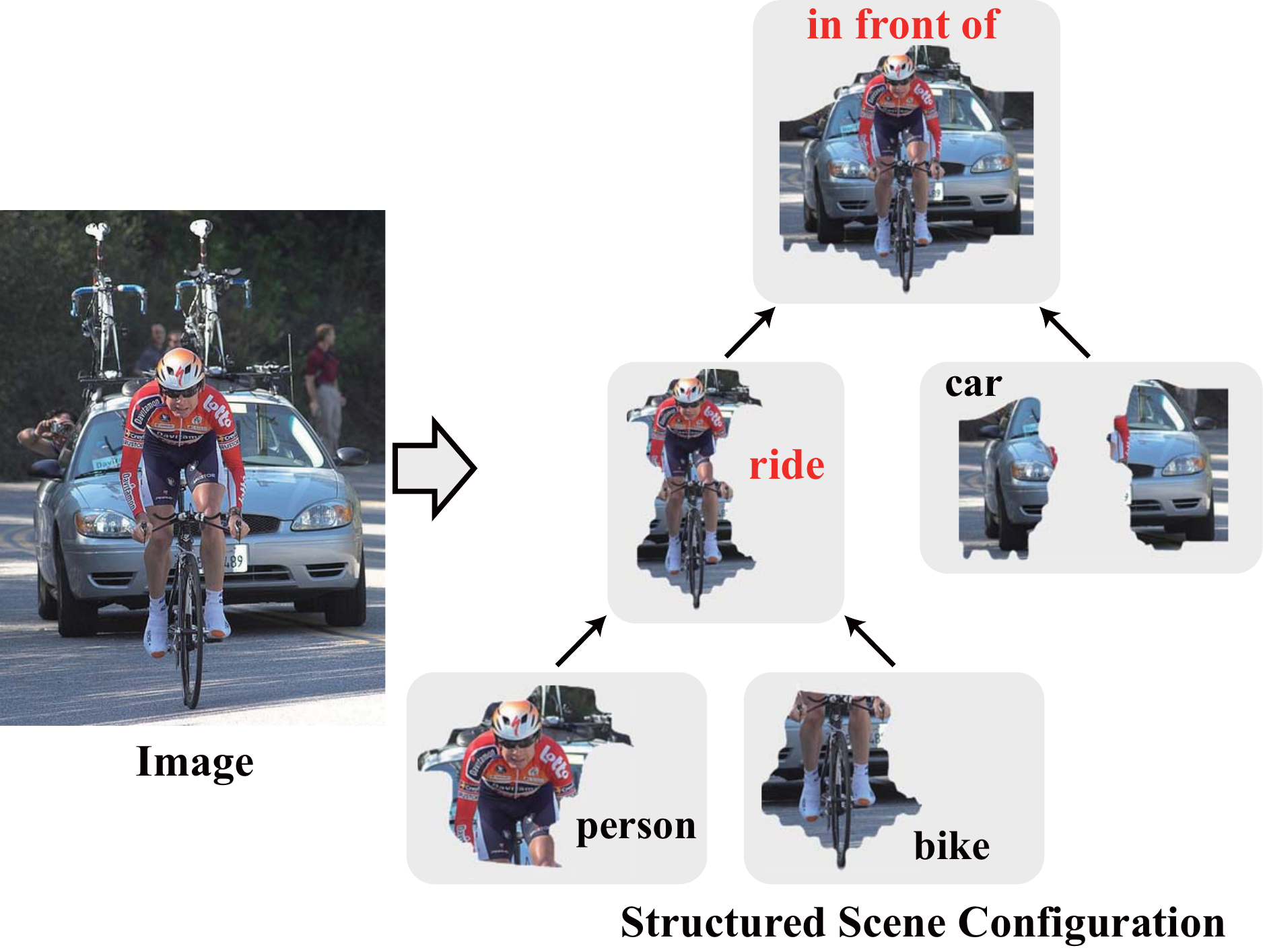}
\caption{An example of structured scene parsing generated by our framework. An input scene image is automatically parsed into a structured configuration that comprises hierarchical semantic objects (black labels) and the interaction relations (red labels) of objects.}
\label{fig:application}
\vspace{-3mm}
\end{figure}

To address these above issues, we develop a novel deep neural network architecture for hierarchical scene parsing. Fig.~\ref{fig:application} shows a parsing result generated by our framework, where a semantic object hierarchy with object interaction relations is automatically parsed from an input scene image. Our model is inspired by the effectiveness of two widely successful deep learning techniques: convolutional neural networks (CNN)~\cite{DBLP:AlexNet}\cite{DBLP:FCnetwork} and recursive neural network (RsNN)~\cite{DBLP:Recursive_Socher}. The former category of models is widely applied for generating powerful feature representations in various vision tasks such as image classification and object detection. Meanwhile, the RsNN models (such as ~\cite{DBLP:Recursive_Socher}\cite{DBLP:RecursiveContext}\cite{DBLP:RecursiveContext2}) have been demonstrated as an effective class of models for predicting hierarchical and compositional structures in image and natural language understanding~\cite{DBLP:RNN-NLP}. One important property of RsNN is the ability to recursively learn the representations in a semantically and structurally coherent way. In our deep CNN-RsNN architecture, the CNN and RsNN models are collaboratively integrated for accomplishing the scene parsing from complementary aspects. We utilize the CNN to extract features from the input scene image and generate the representations of semantic objects. Then, the RsNN is sequentially stacked based on the CNN feature representations, generating the structured configuration of the scene.

On the other hand, to avoid affording the elaborative annotations, we propose to train our CNN-RsNN model by leveraging the image-level descriptive sentences. Our model training approach is partially motivated but different from the recently proposed methods for image-sentence embedding and mapping~\cite{DBLP:VisualAlign-feifei}\cite{DBLP:TellMeShowYou}, since we propose to transfer knowledge from sentence descriptions to discover the scene configurations.

In the initial stage, we decompose each sentence into a semantic tree consisting of nouns and verb phrases with a standard parser~\cite{DBLP:conf/acl/SocherBMN13}, WordNet~\cite{wordnet} and a post-processing method. Then, we develop an Expectation-Maximization-type learning method for model training based on these semantic trees and their associated scene images. Specifically, during the weakly-supervised training, the semantic tree facilitators discover the latent scene configuration in the two following aspects: 1) the objects (\textit{i.e.}, nouns) determine the object category labels existing in the scene, and 2) the relations (\textit{i.e.}, verb phrases) among the entities help produce the scene hierarchy and object interactions. Thus, the learning algorithm iterates in three steps. (i) Based on the object labels extracted from the sentence, it estimates an intermediate label map by inferring the classification probability of each pixel. Multi-scale information of the image is adopted to improve the accuracy. (ii) With the label map, the model groups the pixels into semantic objects and predicts the scene hierarchy and inter-object relations through the RsNN. (iii) With the fixed scene labeling and structure, it updates the parameters of the CNN and RsNN by back propagation.

The main contributions of our work are summarized as follows. i) We present a novel CNN-RsNN framework for generating meaningful and hierarchical scene representations, which helps gain a deeper understanding of the objects in the scene compared with traditional scene labeling.
The integration of CNN and RsNN models can be extended to other high-level computer vision tasks.
ii) We present a EM-type training method by leveraging descriptive sentences that associate with the training images. This method is not only cost-effective but also beneficial to the introduction of rich contexts and semantics.
iii) The advantages of our method are extensively evaluated under challenging scenarios. In particular, on PASCAL VOC 2012, our generated semantic segmentations are more favorable than those by other weakly-supervised scene labeling methods. Moreover, we propose a dedicated dataset for facilitating further research on scene parsing, which contains more than 5000 scene images of 33 categories with elaborative annotations for semantic object label maps, scene hierarchy and inter-object relations.

The remainder of this paper is organized as follows. Section \ref{sec:Related_Work} provides a brief review of the related work. Then we introduce the CNN-RsNN model in Section  \ref{CNN-RNN Architecture} and follow with the model training algorithm in Section~\ref{sec:training}. The experimental results and comparisons are presented in Section \ref{sec:experiments}. Section~\ref{sec:conclusion} concludes the paper and presents some outlook for  future work.

\section{Related Work}
\label{sec:Related_Work}


Scene understanding has been approached through many recognition tasks such as image classification, object detection, and semantic segmentation.
In current research, a myriad of different methods focus on what general scene type the image shows (classification)~\cite{DBLP:VisualAttr}\cite{DBLP:Multi-Class}\cite{DBLP:fine-grained}, what objects and their locations are in a scene (semantic labeling or segmentation)~\cite{DBLP:Seg-03}\cite{DBLP:Seg-09}\cite{DBLP:Seg-15}\cite{tighe2014scene}. These methods, however, ignore or over simplified the compositional representation of objects and fail to gain a deeper and structured understanding on scene.


Meanwhile, as a higher-level task, structured scene parsing has also attracted much attention. A pioneering work was proposed by Tu et al.~\cite{tu2005image}, in which they mainly focused on faces and texture patterns by a Bayesian inference framework. In~\cite{DBLP:ImageParsing-Attribute}, Han et al. proposed to hierarchically parse the indoor scene images by developing a generative grammar model. An extended study also explored the more complex outdoor environment in~\cite{DBLP:3Dparsing-liu}. A hierarchical model was proposed in~\cite{zhu2012recursive} to represent the image recursively by contextualized templates at multiple scales, and rapid inference was realized based on dynamic programming. Ahuja et al.~\cite{ahuja2008connected} developed a connected segmentation tree for object and scene parsing.  Some other related works~\cite{silberman2012indoor}\cite{gupta2013perceptual} investigated the approaches for RGB-D scene understanding, and achieved impressive results. Among these works, the hierarchical space tiling (HST) proposed by Wang et al.~\cite{J:Space-Tiling}, which was applied to quantize the huge and continuous scene configuration space, seemed to be the most related one to ours. It adopted the weakly supervised learning associated the text (\textit{i.e.} nouns and adjectives) to optimize the structure of the parsing graph. But the authors didn't introduce the relations between objects into their method. In terms of the model, HST used a quantized grammar, rather than the neural networks which can adopt the transfer learning to obtain better initialization for higher training efficiency.


With the resurgence of neural network models, the performances of scene understanding have been improved substantially. The representative works, the fully convolutional network (FCN) \cite{DBLP:FCnetwork}  and its extensions \cite{DBLP:CNN-CRF}, have demonstrated effectiveness in pixel-wise scene labeling. A recurrent neural network model was proposed in \cite{DBLP:CRF-RNN}, which improved the segmentation performance by incorporating the mean-field approximate inference, and similar idea was also explored in \cite{DBLP:CNN-MRF}. For the problem of structured scene parsing, recursive neural network (RsNN) was studied in \cite{DBLP:Recursive_Socher}\cite{DBLP:RecursiveContext2}. For example, Socher et al.~\cite{DBLP:Recursive_Socher} proposed to predict hierarchical scene structures with a max-margin RsNN model. Inspired by this work, Sharma et al. proposed the deep recursive context propagation network (RCPN) in~\cite{DBLP:RecursiveContext} and~\cite{DBLP:RecursiveContext2}. This deep feed-forward neural network utilizes the contextual information from the entire image to update the feature representation of each superpixel to achieve better classification performance. The differences between these existing RsNN-based parsing models and our model are three folds. First, they mainly focused on parsing the semantic entities (e.g., buildings, bikes, trees), while the scene configurations generated by our method include not only the objects but also the interaction relations of objects. Second, we introduce a novel objective function to discover the scene structure. Third, we incorporate convolutional feature learning into our deep model for joint optimization.


\begin{figure*}[t!]
\centering
\includegraphics[width=2\columnwidth]{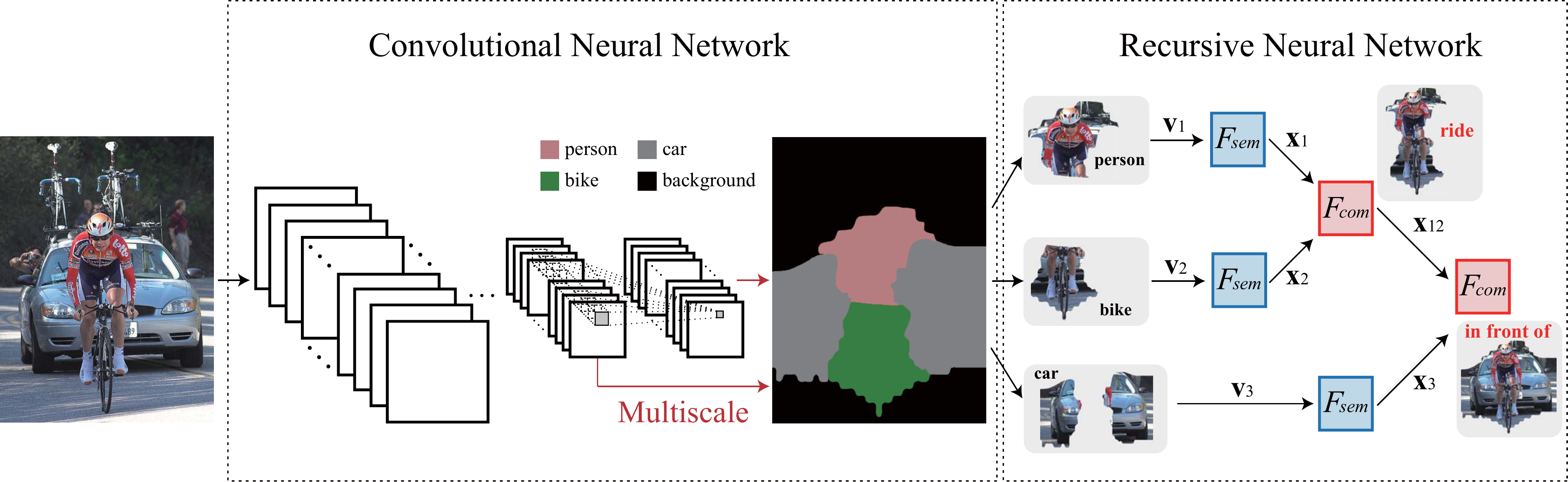}
\caption{The proposed CNN-RsNN architecture for structured scene parsing. The input image is directly fed into the CNN to produce score feature representation of each pixel and map of each semantic category. Then the model applies score maps to classify the pixels, and groups pixels with same labels to obtain  feature representation $\textbf{v}$ of objects. After that $\textbf{v}$ is fed into the RsNN, it is first mapped onto a transition space and then is used to predict the tree structure and relations between objects. $\textbf{x}$ denotes the mapped feature.}
\label{fig:inference}
\vspace{-4mm}
\end{figure*}

Most of the existing scene labeling / parsing models are studied in the context of supervised learning, and they rely on expensive annotations. To overcome this issue, one can develop alternative methods that train the models from weakly annotated training data, e.g., image-level tags and contexts \cite{DBLP:Weakly2}\cite{DBLP:Weakly-MultiInstance}\cite{DBLP:WeaklySegmentation}\cite{DBLP:CCNN}. Among these methods, the one that inspires us is \cite{DBLP:WeaklySegmentation}, which adopted an EM learning algorithm for training the model with image-level semantic labels. This algorithm alternated between predicting the latent pixel labels subject to the weak annotation constraints and optimizing the neural network parameters. Different from this method, our model applies the sentence description to label the salient semantic object in the image. By employing such  knowledge transfer, the model can deal with object labeling and relation prediction simultaneously according to human perception.

\vspace{-3mm}
\section{CNN-RsNN Architecture}
\label{CNN-RNN Architecture}

This work aims to jointly solve three tasks: semantic labeling, scene structure generation, and the inter-object relation prediction. To achieve these goals, we propose a novel deep CNN-RsNN architecture. The CNN model is introduced to perform semantic segmentation by assigning an entity label (\textit{i.e.} object category) to each pixel, and the RsNN model is introduced to discover hierarchical structure and interaction relations among entities.



Fig. \ref{fig:inference} illustrates the proposed CNN-RsNN architecture for structured scene parsing. First, the input image $\mathbf{I}$ is directly fed into revised VGG-16 network~\cite{vggnet} to produce different levels of feature maps. According to these feature maps, multi-scale prediction streams are combined to produce final score maps $\mathcal{S} = \{\mathbf{s}^0, ..., \mathbf{s}^k, ..., \mathbf{s}^K\}$ for object categories. Based on the softmax normalization of score maps,  the $j$-th pixel is assigned with an object label $c_j$. We further group the pixels with the same label into an object, and obtain the feature representations of objects. By feeding these feature representations of objects to the RsNN, a greedy aggregation procedure is implemented for constructing the parsing tree $\mathcal{P}_I$. In each recursive iteration, two input objects (denoted by the child nodes) are merged into a higher-level object
(denoted by the parent node), and generated root note
represents the whole scene. Different from the RsNN architecture in~\cite{DBLP:Recursive_Socher}\cite{DBLP:RecursiveContext2}, our model also predicts the relation between two objects when they are combined into a higher-level object. Please refer to Fig.~\ref{fig:inference} for more information about the proposed architecture. In the following, we discuss the CNN and RsNN models in details.

\subsection{CNN Model}
\label{sub:cnn_model}

The CNN model is designed to accomplish two tasks: semantic labeling and generating feature representations for objects. For semantic labeling, we adopt the fully convolutional network with parameters $\mathbf{W}_C$ to yield $K+1$ score maps $\{\mathbf{s}^0, ..., \mathbf{s}^k, ..., \mathbf{s}^K\}$, corresponding to one extra background category and $K$ object categories. Following the holistically-nested architecture in~\cite{DBLP:HNED-tuzhuowen} we adopt $E=3$ multi-scale prediction streams, and each stream is associated with $K+1$ score maps with the specific scale. Let $s_j^{t,e}$ indicate the score value at pixel $j$ in the $t$-th map of $e$-th scale. We normalize $s_j^{t,e}$ in the $e$-th stream using softmax to obtain the corresponding classification score:
\begin{equation}\label{eq_softmax}
\sigma_e(s_j^{t,e}) = \frac{\exp(s_j^{t,e})} {\sum_{k=0}^K \exp(s_j^{k,e})}
\end{equation}
Then the final classification score $\sigma_f(s_j^{t})$ is further calculated by $\sigma_f(s_j^{t}) = \sum_{e=1}^E \alpha_e \:\sigma_e(s_j^{t,e})$, where $\alpha_e>0 $ is the fusion weight for the $e$-th stream, and $\sum_{e=1}^{E}  \alpha_e = 1$. The learning of this fusion weight is equivalent to training $1\times1$ convolutional filters on the concatenated score maps from all multi-scale streams. $\sigma_f(s_j^t)$ denotes the probability of $j$-th pixel belonging to $t$-th object category with $\sum_{t=1}^K \sigma_f(s_j^t)=1$. The set $\{c_j\}_{j=1}^M$ denotes the predicted labels of pixels in the image $\mathbf{I}$, where $c_j\in \{0,...,K\}$ and $M$ is the number of pixels of image $\mathbf{I}$. With $\sigma_f(s_j^t)$, the label of the $j$-th pixel can be predicted by:

\begin{equation}\label{eq_prediction}
c_j = \arg\max_t \: \sigma_f(s_j^t)
\end{equation}
To generate feature representation for each entity category, we group the pixels with the same label into one semantic category.
%


Considering that the pixel numbers vary with the semantic entity categories, the pooling operation is generally required to obtain fixed-length representation for any object category.
Conventional sum-pooling treats feature representation from different pixels equally, while max-pooling only considers the most representative one and ignores the contribution of the other.
For the tradeoff between sum-pooling and max-pooling, we use \textit{Log-Sum-Exp} (LSE), a convex approximation of the \textit{max} function, as the pooling operator to fuse the features of pixels,
\begin{equation}\label{eq_feature}
\mathbf{v}_k = \frac{1}{\pi}\log \left[\sum_{c_j = k} \exp (\pi \bar{\mathbf{v}}_j) \right]
\end{equation}
where $\mathbf{v}_k$ denotes the feature representation of the $k$-th entity category, $\bar{\mathbf{v}}_j$ denotes the feature representation of the $j$-th pixel by concatenating all feature maps at the layer before softmax at position $j$ into a vector, 
and $\pi$ is a hyper-parameter to control smootheness. One can see that LSE with $\pi=1$ can serve as convex and differentiable approximation of max-pooling~\cite{boyd2004convex}. While LSE with $\pi \rightarrow 0$ degenerates to sum-pooling.


\subsection{RsNN Model}
\label{sub:rnn_model}



\begin{figure}[t]
\centering
\includegraphics[width= 3.0 in]{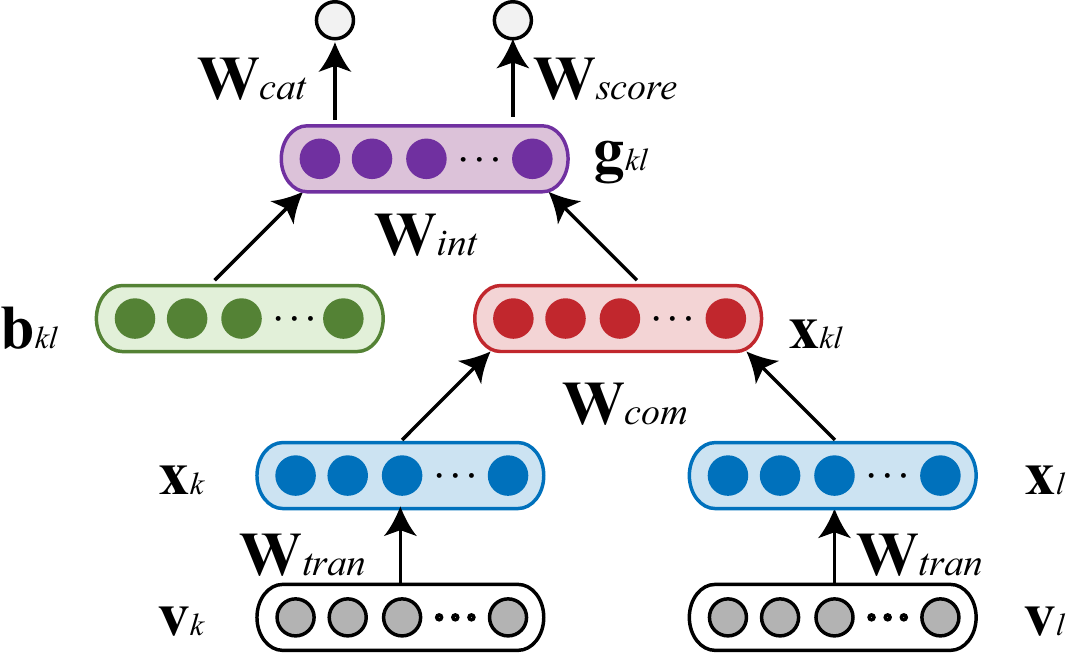}
\caption{An illustration of first layer of proposed recursive neural network which is replicated for each pair of input feature representations. $\mathbf{v}_k$ and $\mathbf{v}_l$ indicate the input feature vectors of two objects. $\mathbf{x}_k$ and $\mathbf{x}_l$ denote the transition features mapped by one-layer fully-connected neural network. The feature representation after the merging operation is denoted by $\mathbf{x}_{kl}$. $\mathbf{W}_{tran}$, $\mathbf{W}_{com}$,$\mathbf{W}_{int}$,$\mathbf{W}_{cat}$ and $\mathbf{W}_{score}$ are parameters of proposed RsNN model. This network is different to the RsNN model proposed in~\cite{DBLP:Recursive_Socher} which only predicts a score for being a correct merging decision. Our model can also be used to predict the interaction relation between the merged objects.}
\label{fig:RNN}
\vspace{-3mm}
\end{figure}

With the feature representations of object categories produced by CNN, the RsNN model is designed to generate the image parsing tree for predicting hierarchical structure and interaction relations. The inputs to scene configuration generation are a set $\Psi$ of nodes, where each node $\mathbf{v}_k \in \Psi$ denotes the feature representation of an object category.
As illustrated in Fig. 3, the RsNN model takes two nodes $\mathbf{v}_k$ and $\mathbf{v}_l$ and their contextual information as the inputs.
The output of RsNN includes three variables: (i) a single real value $h_{kl}$ to denote the confidence score of merging $\mathbf{v}_k$ and $\mathbf{v}_l$, (ii) a relation probability vector $y_{kl}$ for predicting relation label between the two nodes, and (iii) a feature vector $\mathbf{x}_{kl}$ as the combined representation.
In each recursion step, the algorithm considers all pairs of nodes, and choose the pair (e.g., $\mathbf{v}_k$ and $\mathbf{v}_l$) with the highest score to merge.
After the merging, we add $\mathbf{x}_{kl}$ and remove $\mathbf{v}_k$ and $\mathbf{v}_l$ from $\Psi$.
By this way, the nodes are recursively combined to generate the hierarchical scene structure until all the object categories in an image are combined into a root node.

Fig. 3 illustrates the process of RsNN in merging two nodes $\mathbf{v}_k$ and $\mathbf{v}_l$.
In general, the RsNN model is composed of five subnetworks: (i) transition mapper, (ii) combiner, (iii) interpreter, (iv) categorizer, and (v) scorer.
The \emph{transition mapper} is a one-layer fully-connected neural network to generate $\mathbf{x}_k$ and $\mathbf{x}_l$ from $\mathbf{v}_k$ and $\mathbf{v}_l$.
Based on $\mathbf{x}_k$ and $\mathbf{x}_l$, the \emph{combiner} is used to obtain the feature representation $\mathbf{x}_{kl}$.
Then, both $\mathbf{x}_{kl}$ and their contextual information $\mathbf{b}_{kl}$ are considered in the \emph{interpreter} to produce the enhanced feature representation $\mathbf{g}_{kl}$.
Finally, the \emph{categorizer} and \emph{scorer} are used to predict the relation label and confidence score for merging $\mathbf{v}_k$ and $\mathbf{v}_l$.
In the following, we further present more detailed explanation on each subnetwork.

\vspace{3mm}
\textbf{Network Annotations.}
%
Following~\cite{DBLP:Recursive_Socher} and~\cite{DBLP:RecursiveContext2}, object feature $\mathbf{v}_k$ produced by CNN is first mapped onto a transition space by the \textit{\textbf{Transition Mapper}}, which is a one-layer fully-connected neural network.
\begin{equation}\label{eq_sem}
\mathbf{x}_k = F_{tran}(\mathbf{v}_k; \mathbf{W}_{tran})
\end{equation}
where $\mathbf{x}_k$ is the mapped feature, $F_{tran}$ is the network transformation and $\mathbf{W}_{tran}$ indicates the network parameters.  Then the mapped features of two child nodes are fed into the \textit{\textbf{Combiner}} sub-network to generate the feature representation of the parent node.
\begin{equation}\label{eq_com}
\mathbf{x}_{kl} = F_{com}([\mathbf{x}_k, \mathbf{x}_l]; \mathbf{W}_{com})
\end{equation}
where $F_{com}$ is the network transformation and $\mathbf{W}_{com}$ denotes the corresponding parameters. Note that the parent node feature has the same dimensionality as the child node feature, allowing the procedure can be applied recursively.

\begin{figure}[t]
\begin{center}
\includegraphics[width=0.7\linewidth]{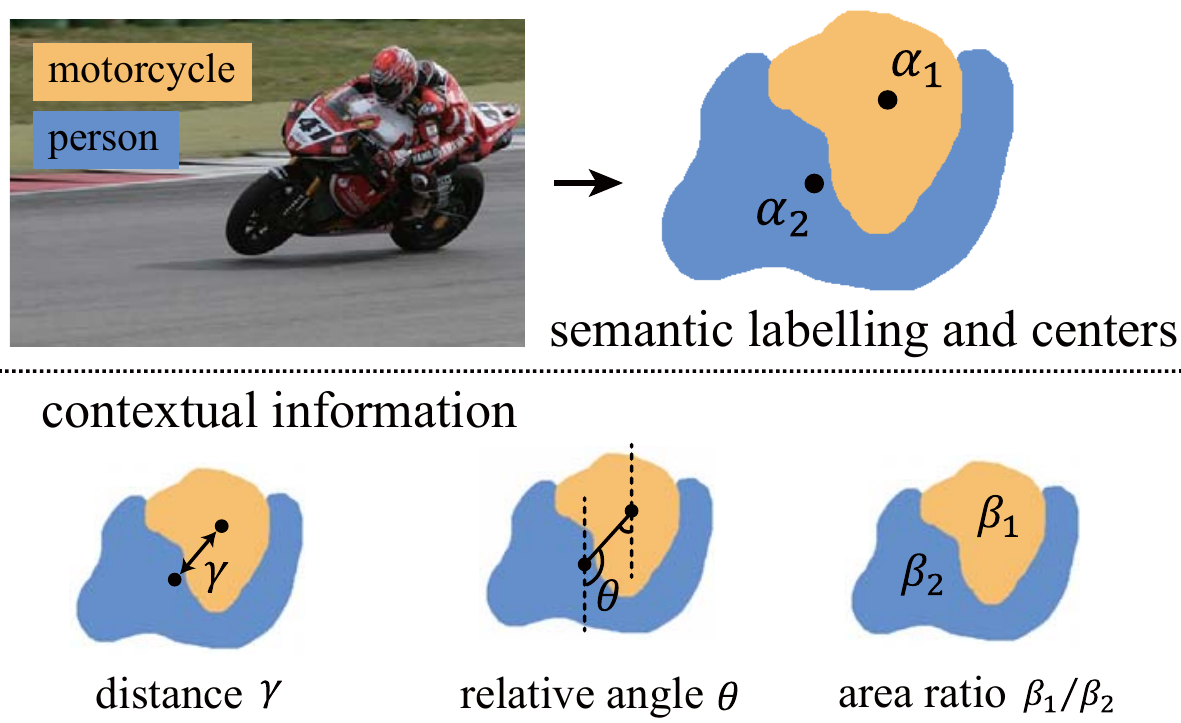}
\end{center}
   \caption{Incorporating the contextual representation into RsNN forward process. The upper row shows the input image and the labeling results of two entities, \textit{i.e}, motorcycle and person. The center of each entity is also given, \textit{i.e.} $\alpha_1$ and $\alpha_2$. Based on the centers and labeling results, the bottom row illustrates three spatial relations, \textit{i.e.}, distance $\gamma$, relative angle $\theta$, and area ratio $\beta_1 / \beta_2$, to characterize the contextual information between the two entities.}
\label{fig:context}
\vspace{-3mm}
\end{figure}

\textit{\textbf{Interpreter}} is the neural network that interprets the relation of two nodes in the parsing tree.
We note that the use of pooling operation in Eqn.~\eqref{eq_feature} will cause the losing of spatial information which is helpful to structure and relation prediction. As a remedy, we design the context features to involve spatial context.
Intuitively, the interpreter network attempts to integrate the feature of two nodes and their contextual information to represent the interaction relation of two entities,
\begin{equation}\label{eq_int}
\mathbf{g}_{kl} = F_{int}([\mathbf{x}_{kl}, \mathbf{b}_{kl}]; \mathbf{W}_{int})
\end{equation}
where $F_{int}$ and $\mathbf{W}_{int}$ indicate the network and layer weights respectively. $\mathbf{b}_{kl}$ denotes the contextual information as follows,
\begin{equation}\label{eq_contextual}
\mathbf{b} = [b^{ang},b^{dis},b^{scal}]
\end{equation}
where $b^{ang}$ and $b^{dis}$ reflect the spatial relation between two semantic entities, while $b^{scal}$ is employed to imply area relation of semantic entities.
As illustrated in Fig.~\ref{fig:context}, $b^{ang}$ denotes the cosine value of angle $\theta$ between the center of two semantic entities. $b^{dis}$ indicates the distance $\gamma$ of two centers (i.e. $\alpha_1$ and $\alpha_2$). $b^{scal}$ is the area rate of such two entities, where $b^{scal} = \beta_1 / \beta_2$. In practice, we normalize all of contextual information into a range of $[-1,1]$.


\textit{\textbf{Categorizer}} sub-network determines the relation of two merged nodes. Categorizer is a softmax classifier that takes relation feature $\mathbf{g}_{kl}$ as input, and predicts the relation label $y_{kl}$,
\begin{equation}\label{eq_rel_category}
y_{kl} = softmax(F_{cat}(\mathbf{g}_{kl}; \mathbf{W}_{cat}))
\end{equation}
where $y_{kl}$ is the predicted relation probability vector, $F_{cat}$ denotes the network transformation and $\mathbf{W}_{cat}$ denotes the network parameters.

\textit{\textbf{Scorer}} sub-network measures the confidence of a merging operation between two nodes. It takes the enhanced feature $\mathbf{g}_{kl}$ as input and outputs a single real value $h_{kl}$.
\vspace{-1mm}
\begin{equation}\label{eq_node_score}
\begin{split}
h_{kl} &= F_{score}(\mathbf{g}_{kl}; \mathbf{W}_{score}) \\
q_{kl} &= \frac{1}{1+exp(-h_{kl})}
\end{split}
\end{equation}
where $F_{score}$ denotes the network transformation and $\mathbf{W}_{score}$ denotes the network parameters. $q_{kl}$ indicates the merging score of node $\{kl\}$. Note such score is important to the configuration discovery and is used to optimize the recursive structure in the training phase, as described in Sec.\ref{sub:rnn_loss}.


\section{Model Training}
\label{sec:training}

Fully supervised training of our CNN-RsNN model requires expensive manual annotations on pixel-level semantic maps, inter-object relations, and hierarchical structure configuration. To reduce the burden on annotations, we present a weakly-supervised learning method to train our CNN-RsNN by leveraging a much cheaper form of annotations, \textit{i.e.}, image-level sentence description. To achieve this goal, the descriptive sentence is first converted to the semantic tree to provide weak annotation information. Then we formulate the overall loss function for structured scene parsing based on the parsing results and the semantic trees. Finally, an Expectation-Maximization (EM) algorithm is developed to train CNN-RsNN by alternatively updating structure configuration and network parameters. In the E-step, guided by the sentence description, we update scene configurations (\textit{i.e.}, intermediate label map $\widehat{\mathbf{C}}$, scene hierarchy and inter-object relations) together with the intermediate CNN and RsNN losses. In the M-step, the model parameters are updated via back-propagation by minimizing the intermediate CNN and RsNN losses.



\subsection{Sentence Preprocessing} \label{sub:preprocessing}

For guiding semantic labeling and scene configuration, we convert each sentence into a semantic tree by using some common techniques in natural language processing. As shown in the bottom of Fig.~\ref{fig:language}, a semantic tree $T$ only includes both entity labels (\textit{i.e.} nouns) and their interaction relations (i.e., verb/ prepositional phrases). Therefore, in sentence preprocessing, we first generate the constituency tree from the descriptive sentence, and then remove the irrelevant leaf nodes and recognize the entities and relations to construct the semantic tree.

The conversion process generally involves four steps. In the first step, we adopt the Stanford Parser~\cite{DBLP:conf/acl/SocherBMN13} to generate the constituency tree (\textit{i.e.} the tree in the top of Fig.~\ref{fig:language}) from the descriptive sentence. Constituency trees are two-way trees with each word in a sentence as a leaf node and can serve as suitable alternative of structured image tree annotation. However, such constituency trees inevitably contain irrelevant words (e.g., adjectives and adverbs) that do not denote semantic entities or interaction relations. Thus, in the second step, we filter the leaf nodes by their part-of-speech, preserving only nouns as object candidates, and verbs and prepositions as relation candidates (\textit{i.e.} the tree in the middle of Fig.~\ref{fig:language}). In the third step, nouns are converted to object categories. Note that sometimes different nouns (e.g. ``cat'' and ``kitten'') represent the same category. The lexical relation in WordNet~\cite{wordnet} is employed to unify the synonyms belonging to the same defined category. The entities that are not in any defined object categories (e.g. ``grass'' in ``a sheep stands on the grass'') are also removed from the trees. In the fourth step, relations are also recognized and refined. Let $\mathcal{R}$ denote a set of defined relations. We provide the list of relations we defined for different datasets in Table~\ref{tbl:relation}. Note that $\mathcal{R}$ also includes an extra relation category, \textit{i.e.} ``others'', to denote all the other relations that are not explicitly defined. Let $\mathcal{T}$ be the set of triplets with the form of $(entity1, verb/prep, entity2)$. We construct a mapping $\mathcal{T} \rightarrow \mathcal{R}$ to recognize the relations and construct the semantic tree (i.e., the tree in the bottom of Fig.~\ref{fig:language}).

\begin{figure*}[t]
\centering
\includegraphics[width=1.0\linewidth]{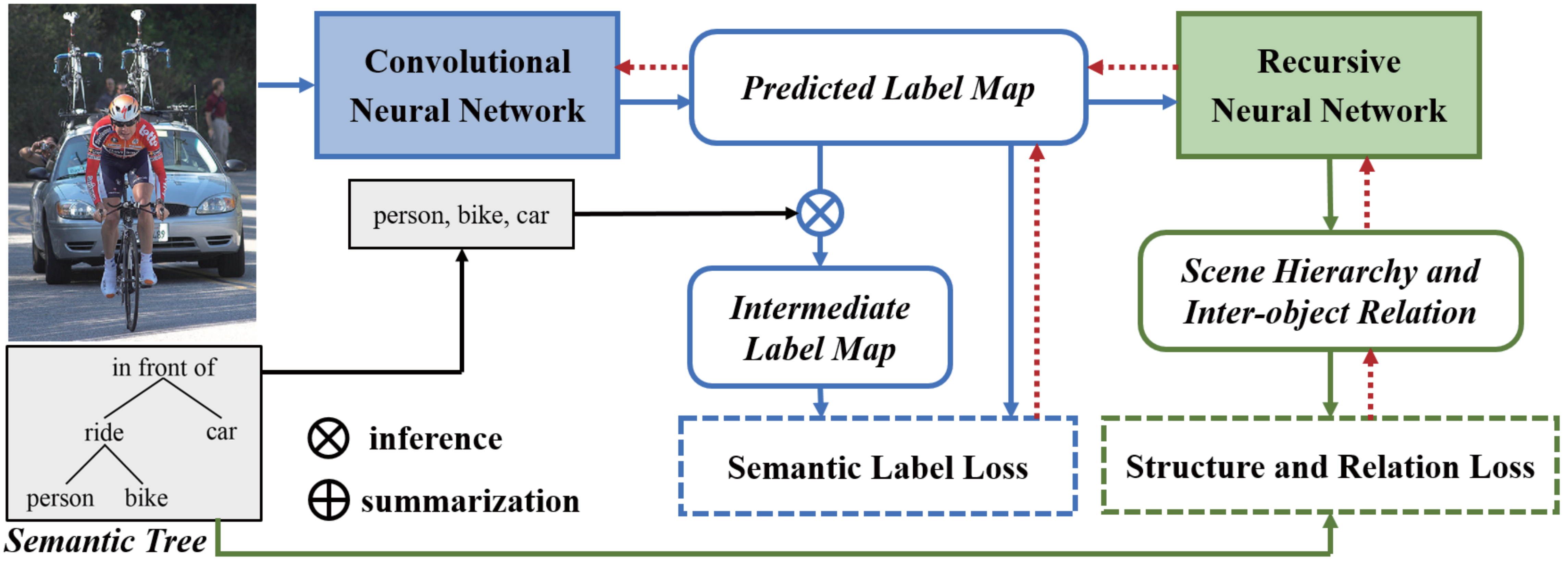}
\vspace{-3mm}
\caption{ An illustration of the training process to our deep model architecture. The blue and green parts are corresponding to semantic labeling and scene structure prediction, respectively. In practice, the input image is first fed into CNN to generate the predicted label map. Then we extract the noun words from the semantic tree to refine the label map, and output intermediate label map. The semantic label loss (\textit{i.e.} the blue dashed block) is calculated by the difference between these two label maps. On the other hand, the feature representation of each object is also passed into RsNN to predict the scene structure. We use scene hierarchy and inter-object relation, and the sematic tree to calculate the structure and relation loss (\textit{i.e.} the green dashed block). The red dotted lines represent the path of back propagation. }
\label{fig:learning}
\end{figure*}

\begin{figure}[t]
\centering
\includegraphics[width=3.5in]{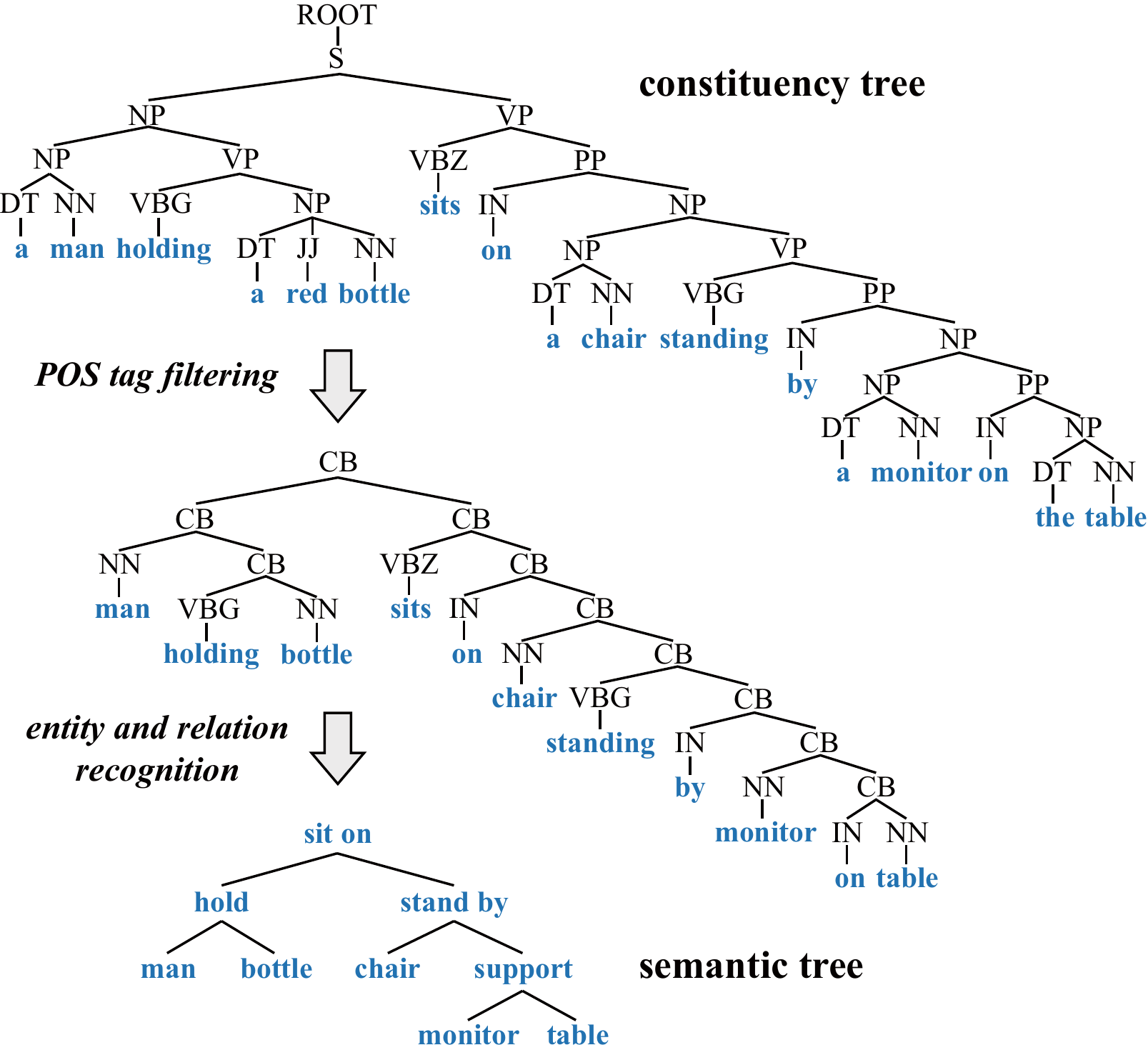}
\caption{An illustration of the tree conversion process. The top is the constituency tree generated by language parser, the middle is the constituency tree after POS tag filtering, and the bottom is the converted semantic tree.}
\label{fig:language}
\end{figure}



\subsection{Loss Functions}

Before introducing the weakly supervised training algorithm, we first define the loss function in the fully supervised setting. For each image $\mathbf{I}_i$, we assume that both the groundtruth semantic map $\mathbf{C}_i$ and the groundtruth semantic tree $T_i$ are known. Then, the loss function is defined as the sum of three terms: semantic label loss $\mathcal{J}_{C}$, scene structure loss $\mathcal{J}_{R}$, and regularizer $R(\mathbf{W})$ on model parameters. With a training set containing $N$ images $\{(\mathbf{I}_1,\mathbf{C}_1, T_1),...,(\mathbf{I}_N, \mathbf{C}_N, T_N)\}$, the overall loss function can be defined as,
\begin{equation}\label{eq_overallloss}
\mathcal{J}(\mathbf{W})  =  \frac{1}{N} \sum_{i=1}^N ( \mathcal{J}_{C}(\mathbf{W}_C;\mathbf{I}_i,\mathbf{C}_i) + \mathcal{J}_{R}(\mathbf{W}; \mathbf{I}_i, T_i) ) + \lambda R(\mathbf{W})
\end{equation}
where $\mathbf{I}_i$ is the $i$-th image. $T_i$ is the groundtruth semantic tree including both hierarchical scene structure and inter-object relation. $\mathbf{W} = \{\mathbf{W}_C, \mathbf{W}_R\}$ denotes all model parameters.
$\mathbf{W}_C$ and $\mathbf{W}_R$ are the model parameters of the CNN and RsNN, respectively.
Note that $\mathbf{W}_R$ includes the parameters of the five subnetworks defined in Sec.3.2, \textit{i.e}. $\mathbf{W}_R = \{ \mathbf{W}_{tran}, \mathbf{W}_{com}, \mathbf{W}_{int}, \mathbf{W}_{cat}, \mathbf{W}_{score} \}$.
The regularization term is defined as $R(\mathbf{W}) = \frac{\lambda}{2}||\mathbf{W}||^2$ and $\lambda$ is the regularization parameter.


\subsubsection{Semantic Label Loss} \label{sub:cnn_loss}

The goal of semantic labeling is to assign the category labels to each pixel.
Let $\mathbf{C}^f$ be the final predicted semantic map, $\mathbf{C}^e$ the $e$-th semantic map of the multi-scale prediction streams. The semantic label loss for an image $\mathbf{I}$ is defined as,
\begin{equation}\label{eq_cnn_loss}
\begin{split}
&\mathcal{J}_{C}(\mathbf{W}_C; \mathbf{I}, \mathbf{C}) = \frac{ \sum_{e=1}^E\mathcal{L}_e(\mathbf{C}, \mathbf{C}^e)}{E} + \mathcal{L}_f(\mathbf{C}, \mathbf{C}^f )
\end{split}
\end{equation}
where $\mathcal{L}_f$ indicates the loss generated by the final predicted semantic map $\mathbf{C}^f$.
Each element in $\mathbf{C}^f$ is calculated by Eqn.~(\ref{eq_softmax}), and we have $\mathbf{C}^{t,f}(j) = \sigma_f(s_j^{t})$. $\mathbf{C}$ is the groundtruth label map. By considering the multi-scale prediction streams, we also define the loss $\mathcal{L}_e, \{e = 1, 2, ..., E\}$ for multiple feature streams (\textit{i.e.} the red line in Fig.~\ref{CNN-RNN Architecture}). Same as the $\mathbf{C}^f$, each element in $\mathbf{C}^e$ is defined by $\mathbf{C}^{t,e}(j) = \sigma_e(s_j^{t,e})$.  The cross entropy is adopted in $\mathcal{L}_f$ and $\mathcal{L}_e$ as the error measure.


\subsubsection{Scene Structure Loss} \label{sub:rnn_loss}

The purpose of constructing scene structure is to generate the meaningful configurations of the scene and predict the interaction relations of the objects in the scene. To achieve this goal, the scene structure loss can be divided into two parts: one for scene hierarchy construction and the other for relation prediction,
\begin{equation}\label{eq_rnn_loss}
	\mathcal{J}_{R}(\mathbf{W}; \mathbf{I}, T) = \mathcal{J}_{struc}(\mathbf{W}_1; \mathbf{I}, T^S) + \mathcal{J}_{rel}(\mathbf{W}_2; \mathbf{I}, T^R)
\end{equation}
where $T^S$ and $T^R$ indicate the groundtruth of hierarchical scene structure and inter-object relations, respectively.
$\mathbf{W}_1 = \{ \mathbf{W}_C, \mathbf{W}_{tran}, \mathbf{W}_{com}, \mathbf{W}_{int}, \mathbf{W}_{score} \}$ and $\mathbf{W}_2 = \{ \mathbf{W}_C, \mathbf{W}_{tran}, \mathbf{W}_{com}, \mathbf{W}_{int}, \mathbf{W}_{cat} \}$.
The above two items are jointly used to optimize the parameters of CNN and RsNN. The difference is that $\mathbf{W}_{score}$ in Eqn.~(\ref{eq_node_score}) and $\mathbf{W}_{cat}$ in Eqn.~(\ref{eq_rel_category}) are optimized by the structure loss and relation loss, respectively.

\textbf{Scene Hierarchy Construction.}
Scene hierarchy construction aims to learn a transformation $\mathbf{I} \rightarrow \mathcal{P}_I$. The predicted scene hierarchy $\mathcal{P}_I$ is said to be valid if the merging order between regions is consistent with that in the groundtruth hierarchical scene structure. Given the groundtruth hierarchical scene structure $T^S$, we extract a sequence of ``correct'' merging operations as $\mathcal{A}(\mathbf{I}, T^S) = \{a_1,...,a_{P_S}\}$, where $P_S$ is the total number of merging operation. Given an operation $a$ on the input image $\mathbf{I}$, we use $q(a)$ to denote the merging score produced by the Scorer sub-network. Based on the merging score $q(a)$ calculated in Eqn.~(\ref{eq_node_score}), we define the loss to encourage the predicted scene hierarchy to be consistent with the groundtruth. Specifically, the score of a correct merging operation is required to be larger than that of any incorrect merging operation $\widehat{a}$ with a constant margin $\triangle$, i.e., $q(a)\geq q(\widehat{a}) + \triangle$. Thus, we define the loss for scene hierarchy construction as,
\vspace{-2mm}
\begin{equation}\label{eq_structure_loss}
\begin{split}
\mathcal{J}_{struc}(\mathbf{W}; \mathbf{I}, T^S) & =  \frac{1}{P_S} \sum_{p=1}^{P_S} [ \:\: \max_{\widehat{a}_p\notin\mathcal{A}(\mathbf{I},T^S)}
q(\widehat{a}_p)\\
& - q(a_p) + \triangle \:]
\end{split}
\end{equation}
Intuitively, this loss intends to maximize the score of correct merging operation while minimizing the scores of incorrect merging operations. To improve efficiency, only the highest score of the incorrect merging operation is considered during training.




\textbf{Relation Categorization.}
Denote by $\{kl\}$ the combination of two child nodes $k$ and $l$. Let $y_{kl}$ be the softmax classification result by the Categorizer sub-network in Eqn.~\eqref{eq_rel_category}, and $\widehat{y}_{kl}$ be the groundtruth relation from $T^R$. The loss on relation categorization is then defined as the cross entropy between $y_{kl}$ and $\widehat{y}_{kl}$,
\begin{equation}\label{eq_relation_loss}
\begin{split}
& \mathcal{J}_{rel}(\mathbf{W}; \mathbf{I}, T^R) = \frac{1}{|N_R|}  \sum_{\{kl\}} \mathcal{L}_r (\widehat{y}_{kl},y_{kl} )
\end{split}
\end{equation}
where $y_{kl}$ is the predicted relation probability in Eqn.~(\ref{eq_node_score}). $|N_R|$ denotes the number of relations in $T^R$.


\begin{small}
\begin{algorithm}[ht]
\caption{EM Method for Weakly Supervised Training}
\label{alg:EM}
\begin{algorithmic}\footnotesize
\REQUIRE ~~\\
    Training samples ($\mathbf{I_1},T_1$),($\mathbf{I_2},T_2$),...,($\mathbf{I_Z},T_Z$).
\ENSURE ~~\\                           
    The parameters of our CNN-RsNN model $\mathbf{W}$
\INPUT ~~\\
    Initialize the CNN model with the pre-trained networks on ImageNet \\
    Initialize the RsNN model with Gaussian distribution
\MYWHILE
\STATE
\begin{itemize}
 \item[1.] Estimate the intermediate semantic maps $\{\widehat{\mathbf{C}}_i\}_{i=1}^Z$ according to Algorithm~\ref{alg:algorithm}
  \item[2.] Predict the scene hierarchy and inter-object relations for each image $\mathbf{I}_i$
  \item[3.] Replace the groundtruth semantic maps $\{\mathbf{C}_i\}_{i=1}^Z$ in Eqn.~(\ref{eq_overallloss}) with intermediate semantic maps $\{\widehat{\mathbf{C}}_i\}_{i=1}^Z$.
  \item[4.] Update the parameters $\mathbf{W}$ according to Eqn.~(\ref{eq_overallloss})
\end{itemize}
\MYENDWHILE \vspace{0.15cm}

The optimization algorithm converges

\end{algorithmic}
\end{algorithm}
\end{small}

\begin{small}
\begin{algorithm}[ht]
\caption{Estimating Intermediate Label Map}
\label{alg:algorithm}
\begin{algorithmic}\footnotesize
\REQUIRE ~~\\
    Annotated entities $T^E$ in the semantic tree, normalized prediction score $\sigma_e(s_j^{k,e})$ and final prediction score $\sigma_f(s_j^{k})$, where $j\in\{1,..,M\}$, $k\in\{0,..,K\}$, $e\in\{1,...,E\}$.
\ENSURE ~~\\                           
    Intermediate label map $\widehat{\mathbf{C}}=\{\widehat{c}_j\}_{j=1}^M$
\INPUT ~~\\
    (1) To simplify, let $f$ be the $E+1$ scale. \\
    (2) Set $\psi^{k,e} = 0$ and $G_j^e(k) = \log \sigma_e(s_j^{k,e})$ for all $e\in \{1,..,E+1\}$ and $k \in \{0,...,K\}$; \\
    (3) Let $\rho_{bg},\rho_{fg}$ indicate the number of pixels being assigned to background and foreground. Set $\rho_k = \rho_{bg}$ if $k=0$, $\rho_k = \rho_{fg}$ if $k\in\{1,...,K\}$.
\MYWHILE
    \STATE
    \begin{itemize}
  \setlength{\itemsep}{1pt}
  \setlength{\parskip}{3pt}
  \setlength{\parsep}{10pt}
  \item[1.] Compute the maximum score at each position $j$, \\ $[G_j^e]_{max} = \max_{k\in T^E} G_j^e(k)$

  \item[2.] \textbf{repeat}

  \textbf{if} the $k$-th semantic category appears in annotated entities $T^E$, \\

  \setlength{\parskip}{1pt}
     \begin{itemize}
     \setlength{\itemsep}{1pt}
     \setlength{\parskip}{3pt}
     \setlength{\parsep}{10pt}
    \: \item[a)] Set $\delta_j^{k,e} = [G_j^e]_{max} - G_j^e(k)$.
       \item[b)] Rank $\{\delta_j^{k,e}\}_{j=1}^M$ according to the ascending sorting and obtain the ranking list.
       \item[c)] Select $\delta_i^{k,e}$ in the $\rho_k$-th position of the ranking list, and let $\psi^{k,e} = \delta_i^{k,e}$
     \end{itemize}
    \textbf{else} Set $\psi^{k,e} = -\infty$ to suppress the labels not present in $T^E$. \vspace{0.01cm} \\

    ~ \\

    Update $G_j^e(k)$ with $G_j^e(k) = \log \sigma_e(s_j^{k,e})+\psi^{k,e}$. \\

    ~ \\

  \textbf{until} Handling all of $K+1$ semantic categories.

  \setlength{\parskip}{2pt}

      \end{itemize}
\MYENDWHILE \vspace{0.15cm}

 Updating all of the prediction score in $E+1$ scales. \\ \vspace{0.15cm}

 Calculate the intermediate label of each pixel using Eqn. (\ref{eq_adaptive})

\end{algorithmic}
\end{algorithm}
\end{small}

\subsection{EM Method for Weakly Supervised Learning} \label{sub:em_learning}


In our weakly supervised learning setting, the only supervision information is the descriptive sentence for each training image. By converting the descriptive sentence to the semantic tree $T$, we can obtain the entities $T^E$ (i.e., nouns), the relations $T^R$ (i.e., verbs or prepositional phrases) and the composite structure $T^S$ between entities, but cannot directly get the semantic map $\mathbf{C}$. Therefore, we treat the semantic labeling map $\mathbf{C}$ as latent variable and adopt a hard EM approximation for model training. In the E-step, we estimate the intermediate semantic map $\widehat{\mathbf{C}}$ based on the previous model parameters and the annotated entities $T^E$, and replace the $\mathbf{C}_i$ in Eqn. (\ref{eq_overallloss}) with its estimate $\widehat{\mathbf{C}}_i$. In the M-step, mini-batch SGD is deployed to update the CNN and RsNN parameters by minimizing the overall loss function. The detail of our EM algorithm is described as follows:

\textbf{(i) Estimate the intermediate semantic map $\widehat{\mathbf{C}}$.} As illustrated in Fig.~\ref{fig:learning} (\textit{i.e.} blue part), the input image $\mathbf{I}$ first goes through the convolutional neural network to generate predicted semantic map. Then the intermediate semantic map $\widehat{\mathbf{C}}$ is estimated based on the predicted map the annotated entities $T^E$,
\begin{equation}\label{eq_adaptive}
\begin{split}
\widehat{\mathbf{C}} = \arg\max_{\mathbf{C}} \log P(\mathbf{C}|\mathbf{I}; \mathbf{W}_C^{'}) + \log P(T^E|\mathbf{C}).
\end{split}
\end{equation}
The classification probability $P(\mathbf{C}|\mathbf{I}; \mathbf{W}_C^{'})$ of each pixel can be computed using Eqn.~\eqref{eq_softmax}. Inspired by the effectiveness of cardinality potentials~\cite{DBLP:conf/uai/TarlowSZAF12}\cite{DBLP:conf/icml/LiZ14}, we define $\log P(T^E|\mathbf{C})$ as entity-dependent bias $\psi^{k}$ for the class label $k$, and set $\psi^{k}$ adaptively in a manner similar to~\cite{DBLP:WeaklySegmentation}.

For multi-scale prediction streams, the score in the $e$-th stream is calculated by $G_j^e(k) = \log \sigma_e(s_j^{k,e}) + \psi^{k,e}$. The fused score is $G_j^f(k) = \log \sigma_f(s_j^{k})+ \psi^{k,f}$. Then the intermediate label of pixel $j$ can be estimated by,
\begin{equation}\label{eq_multifusion}
\begin{split}
\widehat{c}_j=\arg\max_k \: \left[ \: \sum_{e=1}^E G_j^e(k) + G_j^f(k) \: \right]
\end{split}
\end{equation}
Algorithm~\ref{alg:algorithm} summarizes our semantic map estimation method.


\textbf{(ii) Predict the object hierarchy and inter-object relations.} Given the semantic labeling result, we group the pixels into semantic objects and obtain the object feature representations according to Eqn.~(\ref{eq_feature}) in Sec.~\ref{sub:cnn_model}.
Then we use the RsNN model to generate the scene structure recursively.
In each recursion, the model first calculates the context-aware feature representations of two object regions ( object or the combination of objects ) according to Eqn.~(\ref{eq_sem}) $\sim$ Eqn.~(\ref{eq_int}).
Then it merges two object regions with the largest confidence score by Eqn.~(\ref{eq_node_score}) and predict the interaction relation in the merged region by Eqn.~(\ref{eq_rel_category}).
The green part in Fig.~\ref{fig:learning} shows such process.


\textbf{(iii) Update the CNN and RsNN parameters.}
Since the ground truth label map is absent for the weakly supervision manner,
the model applies the intermediated label map estimated in (i) as the pseudo ground truth, and calculates the semantic label loss according to Eqn. \eqref{eq_cnn_loss}.
The blue dashed block in Fig.~\ref{fig:learning} shows this process.
In contract, the structure and relation loss is directly computed by the Eqn. \eqref{eq_rnn_loss}, which uses the semantic tree, scene hierarchy and inter-object relation as the inputs.
The green dashed block in Fig.~\ref{fig:learning} shows such process.
With the mini-batch BP algorithm, the gradients from the semantic label loss propagate backward through all layers of CNN.
The gradients from the scene structure loss first propagate recursively through the layers of RsNN, and then propagate through the object features to the CNN.
Thus, all the parameters (\textit{i.e.}, $\mathbf{W}$) of our CNN-RsNN model can be learned in an end-to-end manner (\textit{i.e.} the red dotted line in Fig.~\ref{fig:learning}).
Algorithm~\ref{alg:EM} summarizes the proposed EM method for weakly supervised training.


\section{Experiments}
\label{sec:experiments}


\begin{table}[t]
\begin{center}
\begin{tabular}{c|c|c|c}

\hline
Method & pixel acc. & mean acc. & mean IoU \\
\hline
\textbf{MIL}-ILP~\cite{pinheiro2015image}&\textbf{71.4}&46.9 & 29.4 \\

\textbf{MIL}-FCN~\cite{DBLP:Weakly-MultiInstance}&69.8&48.2& 28.3 \\

\textbf{DeepLab}-EM-Adapt~\cite{DBLP:WeaklySegmentation}&72.9&52.4&30.3 \\
\hline
\textbf{Ours}-Basic~\cite{DBLP:CNN+RNN} &67.7&56.9& 34.3 \\

\textbf{Ours}-Context & 67.6 &56.9 & 34.4 \\

\textbf{Ours}-MultiScale &68.2&57.4& 34.7 \\

\textbf{Ours}-Full&68.4&\textbf{58.1}& \textbf{35.1} \\
\hline
\end{tabular}
\end{center}
\caption{Results on VOC 2012 \textit{val} set under the weakly supervised learning.}
\label{tbl:result_weak_1}
\end{table}

\begin{table}[t]
\begin{center}
\begin{tabular}{c|c|c|c}

\hline
Method & pixel acc. & mean acc. & mean IoU \\
\hline
\textbf{MIL}-ILP~\cite{pinheiro2015image}&53.1&31.7 &19.9  \\

\textbf{MIL}-FCN~\cite{DBLP:Weakly-MultiInstance}&53.5&31.0& 19.3 \\

\textbf{DeepLab}-EM-Adapt~\cite{DBLP:WeaklySegmentation}&55.9&47.9&20.4 \\
\hline
\textbf{Ours}-Basic~\cite{DBLP:CNN+RNN} &60.1&48.4& 21.5 \\

\textbf{Ours}-MultiScale &60.2&49.2& 21.8\\

\textbf{Ours}-Context &61.1&49.3& 22.5 \\

\textbf{Ours}-Full&\textbf{63.4}&\textbf{49.5}& \textbf{23.7} \\
\hline

\end{tabular}
\end{center}
\caption{Results on SYSU-Scenes under the weakly supervised learning.}
\label{tbl:result_weak_2}
\end{table}



\begin{table}[t]
\begin{center}
\begin{tabular}{c|p{0.7cm}|p{0.6cm}|p{0.5cm}|p{0.5cm}|p{0.5cm}}
\hline
       &       &           & pixel& mean &mean \\
Method & \#strong& \#weak &  acc. & acc. &IoU \\
\hline
\textbf{MIL}-ILP~\cite{pinheiro2015image}&&&\textbf{82.7}& 59.9&39.3 \\

\textbf{MIL}-FCN~\cite{DBLP:Weakly-MultiInstance}&208&1464&82.2&60.3&38.4 \\

\textbf{DeepLab}-EM-Adapt~\cite{DBLP:WeaklySegmentation}&&&81.8&62.6&42.5 \\
\hline

\textbf{Ours}-Basic~\cite{DBLP:CNN+RNN} &&&78.1&62.9& 43.2 \\

\textbf{Ours}-Context &&&78.0&63.4& 43.3 \\

\textbf{Ours}-MultiScale &280&1464&78.2&63.6&43.5  \\

\textbf{Ours}-Full&&&78.2&\textbf{64.1}& \textbf{43.7} \\
\hline
\hline

\textbf{MIL}-ILP~\cite{pinheiro2015image}&&&\textbf{86.4}&65.5& 46.2  \\

\textbf{MIL}-FCN~\cite{DBLP:Weakly-MultiInstance}&1464&1464&86.3&65.7&45.7  \\

\textbf{DeepLab}-EM-Adapt~\cite{DBLP:WeaklySegmentation}&&&85.7&66.6&46.2 \\
\hline

\textbf{Ours}-Basic~\cite{DBLP:CNN+RNN} &&&83.1&70.3& 50.9 \\

\textbf{Ours}-Context &&&83.3&69.9& 51.1 \\

\textbf{Ours}-MultiScale &1464&1464&83.3&70.0& 51.2 \\

\textbf{Ours}-Full&&&83.5&\textbf{70.7}& \textbf{51.7} \\
\hline
\end{tabular}
\end{center}
\caption{Results on VOC 2012 \textit{val} set by ours and other semi-supervised semantic segmentation methods.}
\label{tbl:result_semi_1}
\end{table}

\begin{table}[t]
\begin{center}
\begin{tabular}{c|p{0.7cm}|p{0.6cm}|p{0.5cm}|p{0.5cm}|p{0.5cm}}
\hline
       &       &           & pixel& mean &mean \\
Method & \#strong& \#weak &  acc. & acc. &IoU \\
\hline
\textbf{MIL}-ILP~\cite{pinheiro2015image}&&&59.1&54.3 &27.9  \\

\textbf{MIL}-FCN~\cite{DBLP:Weakly-MultiInstance}&500&2552&53.2&58.1&27.3  \\

\textbf{DeepLab}-EM-Adapt~\cite{DBLP:WeaklySegmentation}&&&60.9&56.8&28.4 \\
\hline

\textbf{Ours}-Basic~\cite{DBLP:CNN+RNN} &&&62.8&57.2&28.8  \\

\textbf{Ours}-Context&&&62.1&57.4&28.9  \\

\textbf{Ours}-MultiScale &500&2552&63.4&\textbf{58.5}&29.6  \\

\textbf{Ours}-Full&&&\textbf{64.4}&57.6&\textbf{29.7}  \\
\hline
\hline

\textbf{MIL}-ILP~\cite{pinheiro2015image}&&&67.8&49.4 & 32.3 \\

\textbf{MIL}-FCN~\cite{DBLP:Weakly-MultiInstance}&1241&2552&67.5&50.9&31.7  \\

\textbf{DeepLab}-EM-Adapt~\cite{DBLP:WeaklySegmentation}&&&66.0&53.1&32.4 \\
\hline

\textbf{Ours}-Basic~\cite{DBLP:CNN+RNN} &&&67.2&53.4&33.7  \\

\textbf{Ours}-Context &&&67.6&51.9&34.0  \\

\textbf{Ours}-MultiScale &1241&2552&70.0&\textbf{54.8}&34.8  \\

\textbf{Ours}-Full&&&\textbf{70.1}&52.3&\textbf{35.5}  \\
\hline
\end{tabular}
\end{center}
\caption{Results on SYSU-Scenes by ours and other semi-supervised semantic segmentation methods.}
\label{tbl:result_semi_2}
\end{table}


In this section, we first apply our method for semantic scene labeling and compare with existing weakly-supervised learning based methods, and then evaluate the performance of our method to generate scene structures. Extensive empirical studies for component analysis are also presented.


\begin{table*}[t]\renewcommand{\arraystretch}{1.2}
\begin{center}
\resizebox{\textwidth}{!}{%
\begin{tabular}{c|c|c|c|c|c|c|c|c|c|c|c|c|c|c|c|c|c|c|c|c|c|c}

\hline
Method & bkg & aero & bike & bird & boat& bottle& bus & car & cat &chair & cow & table & dog & horse & mbike & person & plant & sheep& sofa & train & tv  & \textbf{mean} \\
\hline
\textbf{MIL}-ILP~\cite{pinheiro2015image}&72.2&31.8&19.6&26.0&27.3&33.4&41.8&48.6&42.8&9.96&24.8&13.7&33.2&21.4&30.7&22.4&22.3&27.1&16.6&33.3&19.3&29.4  \\

\textbf{MIL}-FCN~\cite{DBLP:Weakly-MultiInstance}&69.9&29.7&16.5&23.4&23.5&30.3&40.6&46.5&40.8&11.0&28.5&12.0&32.5&22.9&29.8&22.8&19.9&25.3&17.1&31.2& 20.1& 28.3 \\

\textbf{DeepLab}-EM-Adapt~\cite{DBLP:WeaklySegmentation}&71.8&29.7&17.0&24.2&27.1&32.2&43.5&45.4&38.7&10.9&30.0&21.1&33.6&27.7&32.5&32.2&17.9&24.7&19.2&36.4&19.9&30.3 \\
\hline
\textbf{Ours}-Basic~\cite{DBLP:CNN+RNN} &62.4&40.7&20.1&33.5&31.3&25.2&47.9&47.6&42.9&11.1&40.4&22.9&42.2&40.8&40.6&27.8&19.2&36.3&25.0&42.0&21.4&34.3  \\

\textbf{Ours}-Context &62.4&41.0&20.3&34.6&31.8&25.0&48.3&47.6&41.4&11.0&40.1&21.7&43.1&41.6&41.2&27.2&19.0&36.2&24.9&42.2&21.6 & 34.4 \\

\textbf{Ours}-MultiScale &65.3&41.0&20.2&33.8&31.1&25.1&48.7&47.5&42.8&11.1&40.4&22.7&42.4&41.2&41.3&27.6&20.2&37.5&25.0&42.5& 21.5& 34.7 \\

\textbf{Ours}-Full &68.1&40.8&20.8&33.2&32.1&25.8&47.6&47.1&43.7&12.1&41.5&23.1&41.9&40.8&42.6&27.4&20.3&37.3&24.7&42.6&22.3 & \textbf{35.1} \\
\hline

\end{tabular}}%
\end{center}
\caption{Experimental results (\textbf{IoU}) on VOC 2012 \textit{val} set under the weakly supervised learning.}
\label{tbl:result_semantic_segmentation_1}
\end{table*}

\begin{table*}[t]\renewcommand{\arraystretch}{1.2}
\begin{center}
\resizebox{\textwidth}{!}{%
\begin{tabular}{c|c|c|c|c|c|c|c|c|c|c|c|c|c|c|c|c|c}
\hline
Method & bkg & aero &ball & bench &bike& bird & boat& bottle& bus& building & car & cat &chair & cow & cup & dog  & grass \\
\hline
\textbf{MIL}-ILP~\cite{pinheiro2015image} &41.2&36.7&1.93&18.3&28.6&21.6&14.4&9.71&8.79&46.9&20.5&28.2&1.22&17.7&13.9&30.3 &16.7  \\

\textbf{MIL}-FCN~\cite{DBLP:Weakly-MultiInstance} &39.4&31.7&2.13&14.8&25.9&19.5&11.3&8.15&13.1&43.7&20.6&34.5&1.71&17.7&11.5&32.3 & 16.9 \\

\textbf{DeepLab}-EM-Adapt~\cite{DBLP:WeaklySegmentation}&42.4&31.1&2.72&20.2&21.9&14.8&14.7&10.1&10.5&44.2&22.3&34.7&6.59&20.3&10.5&22.8&18.7\\
\hline
\textbf{Ours}-Basic~\cite{DBLP:CNN+RNN}       &46.2&33.7&3.26&20.5&21.9&15.3&20.0&13.8&11.1&44.8&22.5&34.2&7.55&21.0&8.04&23.3 & 16.3 \\

\textbf{Ours}-Context  &45.6&33.3&3.42&20.2&22.6&17.7&19.9&13.9&10.5&43.5&21.3&34.3&7.91&22.5&8.24&23.3& 18.8\\

\textbf{Ours}-MulitScale  &46.1&38.2&3.48&21.8&25.0&19.7&20.9&13.4&12.1&45.3&22.0&35.8&5.96&23.3&8.25&24.1&18.2 \\

\textbf{Ours}-Full        &48.4&39.1&4.08&23.1&26.8&21.0&20.4&13.7&11.7&47.0&24.7&36.1&5.46&24.8&9.01&25.1& 20.3 \\
\hline
\hline
Method & horse & laptop & mbike  & person & racket& rail & sea & sheep& sky & sofa & street  &table & train & tree & TV &umbrella  & \textbf{mean} \\
\hline

\textbf{MIL}-ILP~\cite{pinheiro2015image} &20.2&28.3&47.7&25.9&5.44&12.1&2.71&14.8&10.9&18.2&10.7&14.0&33.8&6.75&25.8 &26.0& 19.9 \\

\textbf{MIL}-FCN~\cite{DBLP:Weakly-MultiInstance} &21.1&29.2&48.2&27.0&5.71&10.6&3.05&14.3&19.7&11.1&9.43&11.5&27.1&7.33&24.5 &22.1& 19.3 \\

\textbf{DeepLab}-EM-Adapt~\cite{DBLP:WeaklySegmentation}&28.4&26.5&39.5&26.3&7.32&17.4&7.49&16.2&16.9&17.3&19.4&14.3&34.5&11.4&19.1 &23.2& 20.4 \\
\hline
\textbf{Ours}-Basic~\cite{DBLP:CNN+RNN} &32.5&29.8&42.5&24.6&7.23&17.5&6.29&17.0&18.1&17.4&20.9&14.6&36.4&13.0&20.9&26.9&21.5 \\

\textbf{Ours}-Context &33.4&20.1&42.6&26.8&7.91&18.6&7.99&14.3&18.5&18.3&22.5&14.4&36.7&13.0&21.3 &26.7& 21.8 \\

\textbf{Ours}-MultiScale &35.0&28.0&44.7&26.7&7.06&19.0&9.02&13.7&18.4&19.8&22.3&13.6&38.8&13.8&21.9&28.7&22.5  \\

\textbf{Ours}-Full &36.1&30.1&48.7&32.9&8.05&19.3&9.91&14.6&18.4&19.7&23.2&14.1&41.2&15.1&22.1&28.9&\textbf{23.7}  \\
\hline

\end{tabular}}%
\end{center}
\caption{Experimental results (\textbf{IoU}) on SYSU-Scenes under the weakly supervised learning learning.}
\label{tbl:result_semantic_segmentation_2}
\end{table*}



\begin{table*}[t]\renewcommand{\arraystretch}{1.2}
\begin{center}
\resizebox{\textwidth}{!}{%
\begin{tabular}{c|c|c||c|c|c|c|c|c|c|c|c|c|c|c|c|c|c|c|c|c|c|c|c|c}

\hline
Method & \#strong & \#weak & bkg & aero & bike & bird & boat& bottle& bus & car & cat &chair & cow & table & dog & horse & mbike & person & plant & sheep& sofa & train & tv  & \textbf{mean}\\
\hline
MIL-ILP~\cite{pinheiro2015image}&&& 82.4 &35.7&23.5&37.3&30.7&40.9&58.0&61.4&56.9&11.1&32.0&13.9&48.2&39.2&44.3&58.2&19.6&38.9&24.1&40.0&29.1 & 39.3  \\
MIL-FCN~\cite{DBLP:Weakly-MultiInstance}&280&1464&81.9&36.5&22.9&32.8&29.2&39.0&57.5&58.8&57.9&11.6&31.4&13.5&47.1&36.1&43.9&57.1&18.9&37.8&23.1&40.5&29.4&38.4\\

DeepLab-EM-Adapt~\cite{DBLP:WeaklySegmentation}&&&83.0&42.8&22.6&40.61&37.5&36.9&60.6&58.5&60.3&15.1&38.5&26.0&51.8&43.6&47.5&58.4&43.7&34.1&24.9&39.9&26.5&42.5\\
\hline

Ours-Basic~\cite{DBLP:CNN+RNN}  && &74.9&50.6&22.9&45.4&41.9&36.9&53.8&58.3&62.4&13.2&49.0&20.9&54.4&50.4&49.1&56.3&23.2&43.0&28.5&45.5&25.5  & 43.2 \\
Ours-Context  &&   &74.8&50.4&23.1&45.5&41.6&37.4&54.4&58.7&62.5&13.1&49.4&21.1&54.5&50.4&49.2&56.5&22.9&43.5&28.5&45.4&25.7 & 43.3 \\
Ours-MultiScale  &280&1464&75.1&50.9&23.2&45.1&42.2&37.2&55.0&59.0&62.7&13.4&49.1&21.1&54.6&50.4&49.4&56.7&22.9&43.8&28.8&46.2&26.1 & 43.5 \\
Ours-Full  &&  &75.0&51.0&23.7&45.7&42.0&37.2&56.9&59.1&62.9&13.4&48.5&22.0&55.1&50.3&48.9&57.0&24.0&43.7&29.2&45.7&26.5  & \textbf{43.7} \\

\hline
\hline

MIL-ILP~\cite{pinheiro2015image}&&&86.5&53.7&24.9&49.0&45.8&48.3&59.4&68.2&64.0&16.8&37.1&14.2&59.2&46.5&54.8&65.3&27.6&37.8&29.1&47.8&34.5&46.2\\
MIL-FCN~\cite{DBLP:Weakly-MultiInstance}&1464&1464&86.2&51.2&23.8&49.9&45.8&47.8&60.9&67.3&64.9&16.6&33.3&11.2&58.3&45.3&57.2&66.7&26.6&37.7&28.2&48.9&33.4 &45.7  \\

DeepLab-EM-Adapt~\cite{DBLP:WeaklySegmentation}&&&85.2&49.5&21.8&51.4&42.6&45.4&63.8&68.9&66.6&16.1&40.9&23.4&56.5&46.4&54.1&64.9&25.4&36.9&26.3&50.6&32.7 &46.2  \\
\hline

Ours-Basic~\cite{DBLP:CNN+RNN}   &&   &80.7&60.6&25.6&55.6&51.9&44.0&61.7&67.2&70.8&16.2&55.3&24.5&64.8&57.7&58.4&66.1&29.6&47.5&35.4&57.1&38.1 & 50.9 \\
Ours-Context  &&  &80.9&61.6&25.5&55.6&52.5&43.3&61.4&66.8&70.8&16.4&55.6&25.4&64.9&57.6&58.3&65.8&29.3&48.4&36.1&55.8&39.6 & 51.1 \\
Ours-MultiScale  &1464&1464&81.3&61.9&25.6&55.9&52.1&43.7&61.6&67.1&71.1&16.2&56.2&24.3&64.7&58.2&58.5&66.1&29.4&47.5&36.3&56.8&40.0  & 51.2 \\
Ours-Full  &&  &81.8&62.4&25.7&55.6&52.3&44.1&62.4&67.8&71.0&16.3&56.6&24.7&65.0&58.7&58.8&66.2&29.7&47.5&37.0&56.8&40.9  & \textbf{51.7} \\
\hline

\end{tabular}}%
\end{center}
\caption{Experimental results (\textbf{IoU}) on VOC 2012 \textit{val} set under the semi-supervised learning.}
\label{tbl:semi_supervision_1}
\end{table*}

\begin{table*}[t]\renewcommand{\arraystretch}{1.2}
\begin{center}
\resizebox{\textwidth}{!}{%
\begin{tabular}{c|c|c|c|c|c|c|c|c|c|c|c|c|c|c|c|c|c|c|c}
\hline
Method &\#strong & \#weak& bkg & aero  &ball &bench &bike & bird & boat& bottle& building &bus & car & cat &chair & cow & cup &dog  & grass \\
\hline
MIL-ILP~\cite{pinheiro2015image} &&&59.8&39.2&3.51&27.5&34.1&20.6&21.6&14.1&17.7&56.6&31.0&25.9&5.58 &27.6& 21.2&39.4&17.9 \\

MIL-FCN~\cite{DBLP:Weakly-MultiInstance} &500&2552&56.0&38.1&3.16&24.6&32.0&27.2&20.1&14.9&15.5&54.7&30.3&24.8&7.91 &27.8&21.5&39.5&10.8  \\

DeepLab-EM-Adapt~\cite{DBLP:WeaklySegmentation}&&&56.9&38.2&6.85&24.1&31.6&18.9&24.0&13.1&13.2&63.9&30.7&41.6&10.0&26.9&14.7 &30.4&29.0  \\
\hline

Ours-Basic~\cite{DBLP:CNN+RNN} &&&56.8&44.8&4.34&26.5&35.7&22.9&23.0&21.4&10.3&57.7&30.6 &37.1&8.09&32.1&13.7&30.5&20.9  \\

Ours-Context &&&56.0&44.1&4.06&26.9&36.0&23.1&24.0&21.1&10.9&58.1&30.3&37.1&7.89&32.4&13.2&29.8&21.6 \\

Ours-MultiScale &500&2552&58.9&42.1&4.76&27.8&36.2&24.1&22.6&21.4&11.7&57.8&31.2&38.3&8.74&32.6&13.4&31.0&22.8  \\

Ours-Full &&&59.8&42.5&4.06&30.7&39.7&23.4&24.4&21.9&11.3&58.0&32.4&38.5&8.42&32.2&12.1&30.2&22.0  \\

\hline
\hline

MIL-ILP~\cite{pinheiro2015image} &&& 67.9&48.6&1.02&29.8&38.4&35.9&19.9&18.4&18.2&54.9&33.4&39.8&6.14 &31.2& 10.7&50.8&36.6 \\

MIL-FCN~\cite{DBLP:Weakly-MultiInstance} &1241&2552&66.0&47.1&1.13&27.5&40.0&37.2&20.1&17.9&19.4&54.1&33.2 &34.8&6.91&30.9&11.5&49.4&38.7  \\

DeepLab-EM-Adapt~\cite{DBLP:WeaklySegmentation}&&&61.5&48.9&7.73&26.8&40.7&22.9&27.3&13.8&12.6&56.8&36.9&46.7&11.6 &45.2&17.4&44.3&30.9  \\
\hline

Ours-Basic~\cite{DBLP:CNN+RNN}  &&&62.3&51.5&7.01&28.9&41.6&30.1&30.7&25.8&21.7&53.4&36.0&40.5&9.71&47.0&17.6&49.8&26.1  \\

Ours-Context &&&65.0&50.7&7.13&28.1&41.2&36.9&25.3&24.9&11.8&53.1&35.9&42.3&9.48&46.9&17.2&49.0&30.7 \\

Ours-MultiScale &1241&2552&62.3&54.4&7.15&27.7&43.8&32.8&31.6&23.5&16.6&57.2&36.7&40.4&11.1&43.8&19.2&48.3&32.1 \\

Ours-Full  &&&65.1&54.5&9.19&30.1&43.6&36.8&27.9&23.3&16.3&58.4&38.1&38.8&11.6&47.4&20.8&48.8&33.1 \\

\bottomrule
\toprule

Method &\#strong & \#weak & horse & laptop & mbike & person & racket& rail &sea & sheep& sky & sofa  & street &table & train & tree & TV& umbrella  & \textbf{mean} \\
\hline

MIL-ILP~\cite{pinheiro2015image} &&&39.5&41.5&45.5&45.0&12.8&15.2&20.3&35.8&15.1&25.9&24.7&17.6&43.1&17.4&21.1&37.8& 27.9  \\

MIL-FCN~\cite{DBLP:Weakly-MultiInstance}&500&2552&37.6&39.7&47.2&41.5&12.0&15.1&21.1&34.5&18.8&28.3&26.7 &17.7& 42.8&19.1&19.7&31.5& 27.3 \\

DeepLab-EM-Adapt~\cite{DBLP:WeaklySegmentation}& & &36.3&34.1&50.1&41.0&8.06&23.9&15.9&39.2&25.2&20.8&32.3&14.7&48.2 &16.2& 25.4& 30.6 &28.4 \\
\hline

Ours-Basic~\cite{DBLP:CNN+RNN} &&&37.0&35.6&53.4&45.6&8.04&13.7&13.8&38.4&26.3&20.5&29.8 &19.1&47.5 &21.2&26.6&36.8& 28.9\\

Ours-Context &&&36.6&36.3&51.3&44.4&8.64&19.6&11.7&36.8&24.7&20.3&29.7 &20.4&48.8 &21.1&26.2&36.9& 28.9 \\

Ours-MultiScale &500&2552&38.3&39.1&54.9&48.7&8.45&20.8&10.4&37.6&27.0&21.2&30.6&19.4&49.9&19.3&27.0&37.4& 29.4\\

Ours-Full  &&&40.3&37.6&55.8&55.0&8.91&12.7&15.6&37.7&26.8&21.8&29.9&19.8&48.7&18.8&23.1&37.9& \textbf{29.7}\\

\hline
\hline

MIL-ILP~\cite{pinheiro2015image} &&&43.5&39.6&53.7&56.2&15.2&13.2&21.4&24.7&37.0&29.7&23.2 &18.8&40.6&23.6&35.5&48.7&32.3  \\

MIL-FCN~\cite{DBLP:Weakly-MultiInstance} &1241&2552&47.6&39.7&55.2&51.5&12.2&14.1&21.5&24.5&35.8&28.3&16.7 &17.7& 42.8&21.0&29.7&51.5&31.7 \\

DeepLab-EM-Adapt~\cite{DBLP:WeaklySegmentation}&&&42.1&41.1&56.5&49.4&10.6&14.5&26.2&29.8&29.9&31.0&36.7&17.9&49.0 &14.7&30.5& 39.4&32.4 \\
\hline

Ours-Basic~\cite{DBLP:CNN+RNN} &&&51.2&43.5&57.8&56.6&9.83&8.44&28.5&28.0&28.3&33.7&27.8&20.1&50.5&18.5&26.7&41.6&33.7  \\

Ours-Context &&&52.4&42.5&58.8&57.5&10.7&12.3&31.9&28.3&24.9&32.9&33.2&20.8&51.3&16.3&27.0&46.6 &34.0\\

Ours-MultiScale &1241&2552&51.5&42.2&58.6&62.3&11.0&10.0&34.8&30.3&29.8&33.8&25.6&22.6&51.7&24.5&30.2&43.2& 34.8 \\

Ours-Full  &&&51.5&44.3&60.9&62.0&11.1&12.6&36.2&29.1&32.1&34.1&23.2&21.6&52.8&24.4&30.0&44.2&\textbf{35.5}  \\
\hline

\end{tabular}}%
\end{center}
\caption{Experimental results (\textbf{IoU}) on SYSU-Scenes under the semi-supervised learning learning.}
\label{tbl:semi_supervision_2}
\end{table*}


%


\begin{figure*}[t]
\centering
\includegraphics[width=6.5in]{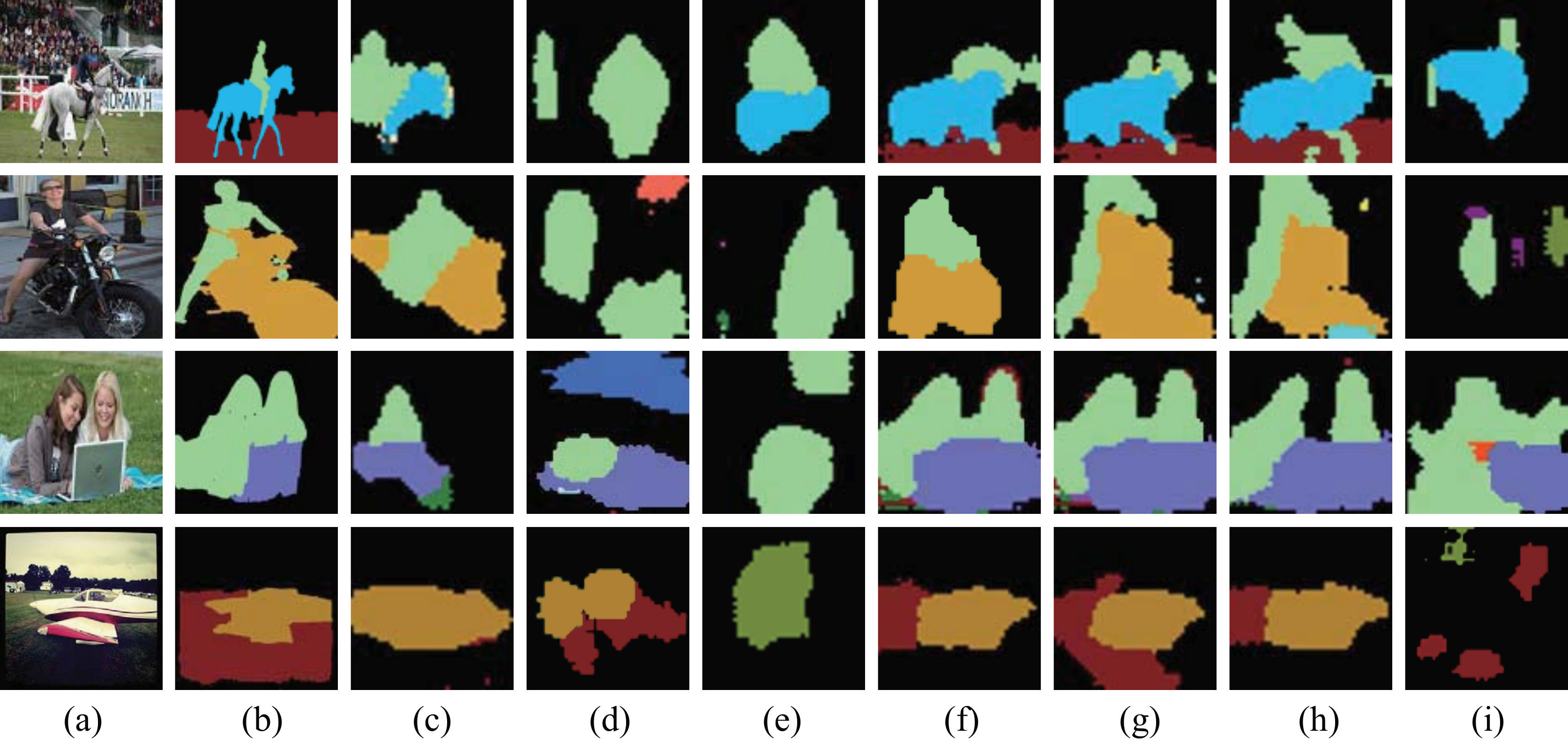}
\vspace{-4mm}
\caption{Visualized semantic labeling results on SYSU-Scenes. (a) The input images; (b) The groundtruth labeling results; (c) Our proposed method (weakly-supervised); (d) DeepLab (weakly-supervised)~\cite{DBLP:WeaklySegmentation}; (e) MIL-ILP (weakly-supervised)~\cite{pinheiro2015image}; (f) Our proposed method (semi-supervised with 500 strong training samples); (g) Our proposed method (semi-supervised with 1241 strong training samples); (h) DeepLab(semi-supervised with 500 strong training samples)~\cite{DBLP:WeaklySegmentation}; (i) MIL-ILP (semi-supervised with 500 strong training samples)~\cite{pinheiro2015image}.}.
\label{fig:labelling_result_2}
\end{figure*}

\begin{figure*}[t] \centering
\includegraphics[width= 7 in]{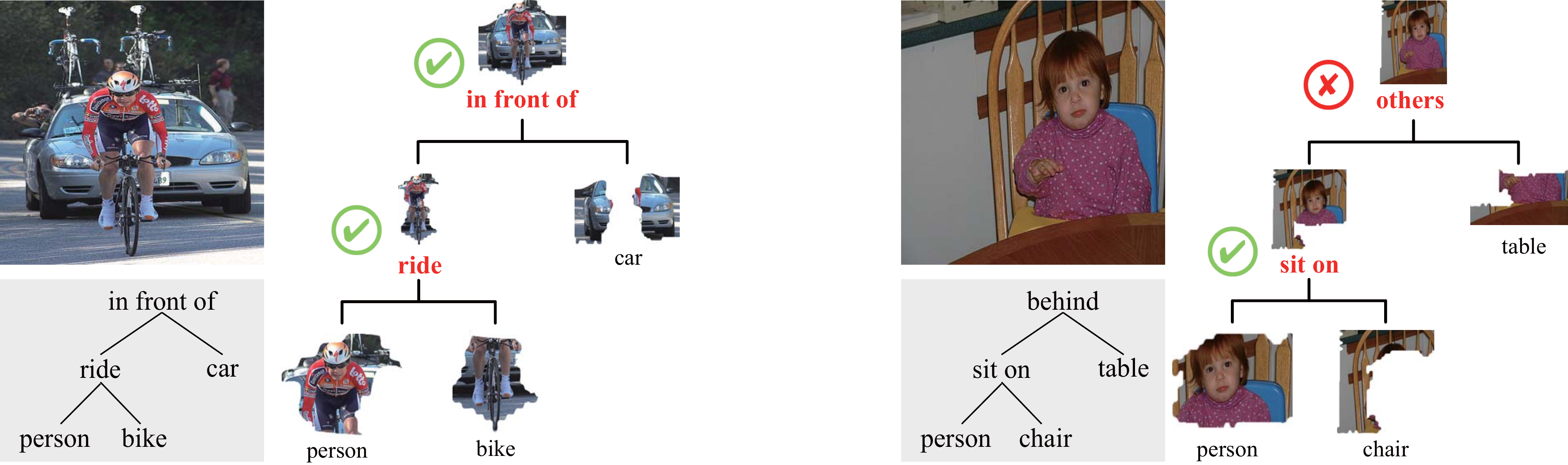}
\vspace{-2mm}
\caption{Visualized scene parsing results on PASCAL VOC 2012 under the weakly-supervised setting. The left one is a successful case, and the right is a failure one. In each case, the tree on the left is produced from descriptive sentence, and the tree on the right is predicted by our method.}
\label{fig:relation1}
\end{figure*}

\begin{figure*}[t] \centering
\includegraphics[width= 7 in]{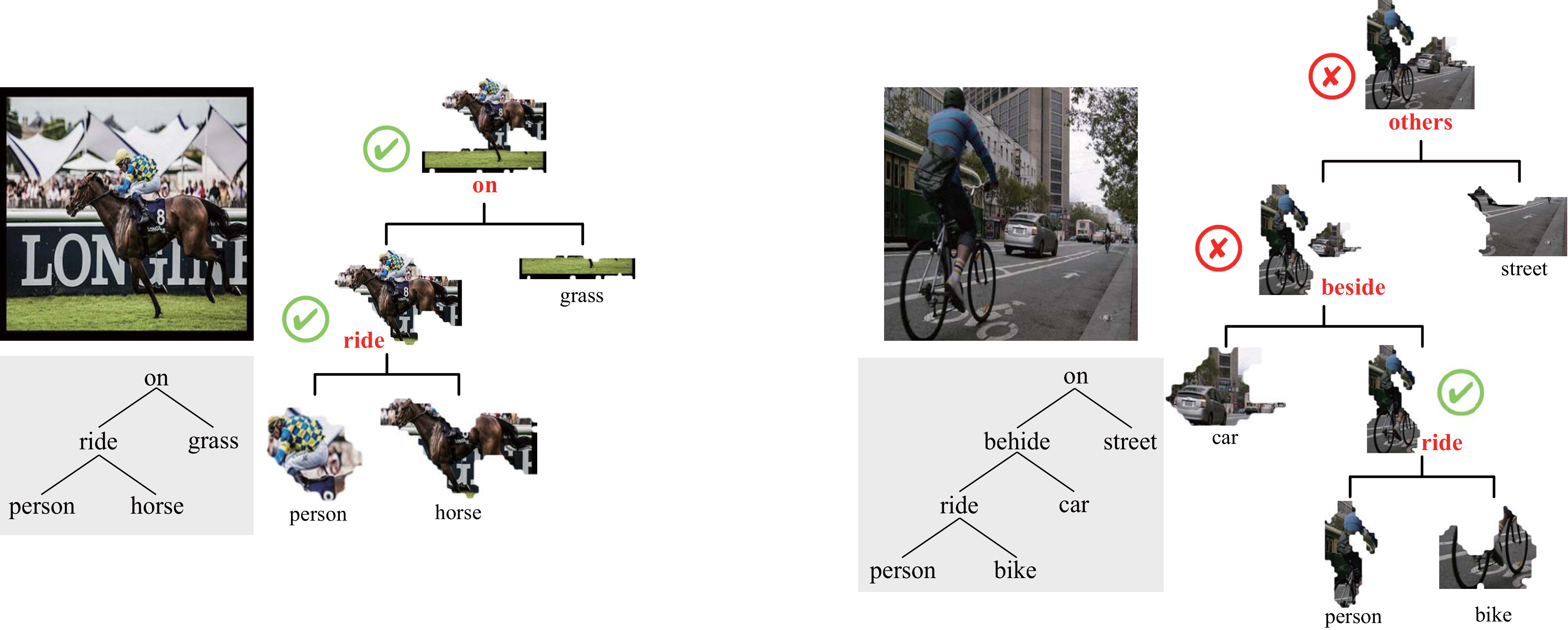}
\vspace{-3mm}
\caption{Visualized scene parsing results on SYSU-Scenes under the semi-supervised setting (i.e. with 500 strongly-annotated images). The left one is a successful case, and the right is a failure one. In each case, the tree on the left is produced from descriptive sentence, and the tree on the right is predicted by our method.}
\label{fig:relation2}
\end{figure*}


\begin{table}[t]
\begin{center}
\begin{tabular}{c|c|c|c}

\hline
Method & \# strong & \# weak  & mean IoU \\
\hline
\textbf{MIL}-ILP~\cite{pinheiro2015image}&  &  & 28.81\\

\textbf{DeepLab}-EM-Adapt~\cite{DBLP:WeaklySegmentation}& 0 & 1464 & 30.57\\

\textbf{Ours}-Full& &  & 35.19\\

\hline

\textbf{MIL}-ILP~\cite{pinheiro2015image}&  &  & 39.03\\

\textbf{DeepLab}-EM-Adapt~\cite{DBLP:WeaklySegmentation}& 280 & 1464 &43.07\\

\textbf{Ours}-Full& &  & 44.13\\

\hline

\textbf{MIL}-ILP~\cite{pinheiro2015image}&  &  & 46.14\\

\textbf{DeepLab}-EM-Adapt~\cite{DBLP:WeaklySegmentation}& 1464 & 1464 & 46.82\\

\textbf{Ours}-Full& &  &51.37\\

\hline

\end{tabular}
\end{center}
\caption{Performance on PASCAL VOC 2012 \textit{test} set.}
\label{tbl:VOC_TEST}
\end{table}


\subsection{Experimental Setting}
\label{sub:experimental_setup}

\textbf{Datasets.}
We adopt \textbf{PASCAL VOC 2012} segmentation benchmark \cite{pascal_voc} in our experiments, which includes 20 foreground categories and one background category. And 1,464 annotated images are used for training and 1,449 images for validation. Note that we exclude the original testing subset on this benchmark due to the lack of available ground-truth annotations.

We also introduce a new dataset created by us, i.e., \textbf{SYSU-Scenes}\footnote{http://www.sysu-hcp.net/SYSU-Scenes/}, especially for facilitating research on structured scene parsing. {SYSU-Scenes} contains 5,046 images in 33 semantic categories, in which 3,000 images are selected from Visual Genome dataset~\cite{VisualGenome} and the rest are crawled from Google. For each image, we provide the annotations including semantic object label maps, scene structures and inter-object relations. We divide the dataset into a training set of 3,793 images and a test set of 1,253 images. Compared with existing scene labeling / parsing datasets, SYSU-Scenes includes more semantic categories (i.e., 33), detailed annotations for scene understanding, and more challenging scenarios (e.g., ambiguous inter-object relations and large intra-class variations).



\textbf{Sentence Annotation.}
We annotate one sentence description for each image in both PASCAL VOC 2012 and SYSU-Scenes. Since our work aims to learn a CNN-RsNN model for category-level scene parsing and structural configuration, in the supplementary materials, we explain the principles of sentence annotation in more details, and provide representative examples and statistics of the sentence annotation. All the descriptive sentences on the VOC 2012 train and val sets are also given.


The sentence description of an image naturally provides a tree structure to indicate the major objects along with their interaction relations~\cite{DBLP:journals/ml/Elman91}. As introduced in Section~\ref{sub:preprocessing}, we use the Stanford Parser~\cite{DBLP:conf/acl/SocherBMN13} for sentence parsing and further convert the parsing result into the regularized semantic tree. In this work, we see to it that the semantic tree is generated from one sentence.


\begin{table}[t]
\begin{center}
\begin{tabular}{c|c|p{4cm}<{\centering}}

\hline
Dataset & Amount & Relations  \\
\hline
                 &      &\textit{beside, lie, hold, ride, behind,} \\
PASCAL VOC 2012  &  9   &\textit{sit on, in front of, on} and \textit{others.}\\
\hline
                 &      &\textit{behind, beside, fly, hold, play,} \\
SYSU-Scenes     &  13  & \textit{in front of, ride, sit on, stand, under, walk, on,}  and \textit{others}.\\
\hline

\end{tabular}
\end{center}
\caption{The defined relations in PASCAL VOC 2012 and SYSU-Scenes.}
\label{tbl:relation}
\end{table}



\textbf{Network Architecture and Training.}
Our deep architecture is composed of the stacked CNN and RsNN modules using the Caffe~\cite{DBLP:Caffe} framework. We apply the VGG network~\cite{vggnet} to build the CNN module of 16 layers, and the RsNN is implemented by four extra neural layers upon the CNN. Our network thus contains 20 layers.

All models in our experiment are trained and tested on a single NVIDIA Tesla K40. The parameters of the VGG-16 network are pre-trained on ImageNet~\cite{DBLP:AlexNet}, and the other parameters are initialized with Gaussian distribution with standard deviation of 0.001. We train our network using stochastic gradient descent (SGD) with the batch size of 9 images, momentum of 0.9, and weight decay of 0.0005. The learning rate is initialized with 0.001. We train the networks for roughly 15,000 iterations, which takes 8 to 10 hours.

\subsection{Semantic Labeling}
\label{sub:semantic_Labelling}

To evaluate the semantic scene labeling performance of our method, we re-scale the output pixel-wise prediction back to the size of original groundtruth annotations. The indicators, i.e., \textbf{pixel accuracy}, \textbf{mean class accuracy} and \textbf{mean intersection over union (IoU)}~\cite{DBLP:FCnetwork}, are adopted for performance evaluation. We consider two ways of training our CNN-RsNN model, i.e., weakly-supervised learning and semi-supervised learning.


\textbf{Weakly-supervised Learning.}
We compare our method with several state-of-the-art weakly-supervised semantic segmentation approaches, including MIL-ILP~\cite{pinheiro2015image}, MIL-FCN~\cite{DBLP:Weakly-MultiInstance} and DeepLab~\cite{DBLP:WeaklySegmentation}. We perform experiments with the publicly available code of DeepLab, and our own implementation of MIL-ILP and MIL-FCN.
In practice, we extract the multi-class labels of each image from its groundtruth label map as the supervision information to train the competing models. As for our method, we apply the noun words in the semantic trees as the image-level labels.
Table \ref{tbl:result_weak_1} and Table \ref{tbl:result_weak_2} list the results of the three performance metrics on PASCAL VOC 2012 and SYSU-Scenes. Table \ref{tbl:result_semantic_segmentation_1} and Table \ref{tbl:result_semantic_segmentation_2} further report the breakdown IoU results with respect to object category. Our method obtains the mean IoUs of 35.1\% and 23.7\% on the two datasets, outperforming DeepLab\cite{DBLP:WeaklySegmentation} by 4.8\% and 3.3\%, respectively.

\textbf{Semi-supervised Learning.}
Moreover, we evaluate our method under the way of semi-supervised model learning.
In this setting, the groundtruth semantic labeling maps are available for a part of images in the training set, and others still use the image-level category labels as the supervision.
Our CNN-RsNN model can be easily trained on strongly-annotated images without estimating their intermediate label maps. Following the setting of existing semi-supervised learning based methods on PASCAL VOC 2012, we employ part of images from the Semantic Boundaries dataset (SBD)~\cite{sbd_dataset} to conduct the experiments: using $280$ and $1464$ strongly-annotated images from SBD, respectively, in addition to the original $1464$ weakly annotated (i.e., associated sentences) images. We set the weight, i.e., 1 : 1, for combining the loss scores that respectively computed on the strongly-annotated images and weakly-annotated images. Table \ref{tbl:result_semi_1} reports the quantitative results generated by our method and other competing approaches. Table \ref{tbl:semi_supervision_1} presents the breakdown IoU results on each object category. We also conduct the experiments on SYSU-Scenes, and select $500$ and $1241$ images from the training set as the strongly-annotated samples, respectively. And the overall results are reported in Table \ref{tbl:result_semi_2} and the breakdown IoU results in Table \ref{tbl:semi_supervision_2}.

It can be observed that all methods benefit from the strongly-annotated supervision. On PASCAL VOC 2012, compared with our weakly supervised CNN-RsNN baseline, the improvement on IoU is 8.6\% with 280 strongly annotated images (amount of ``strong'' : ``weak'' samples $=$ 1:5), and is 16.6\% with 1464 strongly annotated images (amount of ``strong'' : ``weak'' samples = 1:1). Moreover, our method outperforms semi-supervised DeepLab~\cite{DBLP:WeaklySegmentation} by 1.2\% with 280 strongly-annotated samples and 5.5\% with 1464 strongly-annotated ones. On SYSU-Scenes, in terms of IoU, our model outperforms the weakly-supervised CNN-RsNN baseline by 6.0\% with 500 strongly-annotated images (amount of ``strong'' : ``weak'' samples  $=$ 1:5), and 11.8\% with 1241 strongly annotated images (amount of ``strong'' : ``weak'' samples  $=$ 1:2). Our model also outperforms semi-supervised DeepLab~\cite{DBLP:WeaklySegmentation} by 1.3\% with 500 strongly-annotated images and 3.1\% with 1241 strongly-annotated images. Finally, Fig.~\ref{fig:labelling_result_2} presents the visualized labeling results on SYSU-Scenes.

To follow the standard protocol for PASCAL VOC semantic segmentation evaluation, we also report the performance of our method on the VOC 2012 \textit{test} dataset in Table~\ref{tbl:VOC_TEST}, under both the weakly-supervised and semi-supervised manners.

\subsection{Scene Structure Generation}
\label{sub:structure_semantic_parsing}

Since the problem of scene structure generation is rarely addressed in literatures, we first introduce two metrics for evaluation: \textbf{structure accuracy} and \textbf{mean relation accuracy}. Let $T$ be a semantic tree constructed by CNN-RsNN and $P = \{T, T_1, T_2, \ldots, T_m\}$ be the set of enumerated sub-trees (including $T$) of $T$. A leaf $T_i$ is considered to be correct if it is of the same object category as the one in the ground truth semantic tree. A non-leaf $T_i$ (with two subtrees $T_l$ and $T_r$) is considered to be correct if and only if $T_l$ and $T_r$ are both correct and the relation label is correct as well. Then, the relation accuracy is defined as $\frac{(\# of correct subtrees)}{m+1}$ and can be computed recursively. The mean relation accuracy is the mean of relation accuracies across relation categories.  Note that the number of sub-trees of each relation category is highly imbalanced in both two datasets, where the relations of most sub-trees are from several dominant categories. Taking this factor into account, the mean relation accuracy metric should be more reasonable than the relation accuracy metric used in our previous work~\cite{DBLP:CNN+RNN}.

Here we implement four variants of our CNN-RsNN model for comparison, in order to reveal how the joint learning of CNN-RsNN and the utility of context contribute to the overall performance. To train the CNN-RsNN model, we consider two learning strategies: i) updating all parameters of the RsNN by fixing the parameters of CNN; ii) joint updating the parameters of CNN and RsNN in the whole process. For each strategy, we further evaluate the effect of  contextual information (i.e., distance, relative angle and area ratio) by learning the interpreter sub-networks (i) with contextual information and (ii) without contextual information.




\begin{table}[t]
\begin{center}
\begin{tabular}{c|c|c|c|c}

\hline
&CNN & RsNN & struct. & mean rel.\\

\hline
Without &partial fixed & updated & 61.7 &  27.9 \\

Context &updated & updated & 64.2 &  28.6 \\

\hline

With    &partial fixed & updated & 62.8 & 27.4 \\

Context &updated & updated & \textbf{67.4} & \textbf{32.1}\\
\hline

\end{tabular}
\end{center}
\caption{Results on PASCAL VOC 2012 with different learning strategies.}
\label{tbl:result_end_to_end_learning}
\vspace{-5mm}
\end{table}

\begin{table}[t]
\begin{center}
\begin{tabular}{c|c|c|c|c}

\hline
   &CNN & RsNN & struct. & mean rel.\\
\hline
Without &partial fixed & updated & 38.0 & 19.6\\

Context &updated & updated & 44.3& 24.1 \\

\hline

With    &partial fixed & updated & 41.7 & 21.8\\

Context &updated & updated & \textbf{48.2} &  \textbf{24.5}\\
\hline

\end{tabular}
\end{center}
\caption{Results on SYSU-Scenes with different learning strategies.}
\label{tbl:result_end_to_end_learning_2}
\vspace{-8mm}
\end{table}


\begin{table*}[t]
\begin{center}
\begin{tabular}{c|c|c|c|c|c|c|c|c|c|c|c|c}

\hline
       &CNN & RsNN & beside & lie & hold & ride & behind & sit & in front & on & other  & \textbf{mean} \\
\hline
 Without &partial fixed & updated & 20.7 & 4.54 &3.57 &23.4 &14.3 &81.1 &2.77 &59.0  &34.9 &27.9\\

 Context &updated & updated & 23.6 & 13.6 & 14.3 & 33.3 & 17.8 & 64.8 & 7.93 &46.3 & 35.6 & 28.6 \\

\hline

With      & partial fixed & updated & 18.3 & 18.2 &17.8 &40.7 &10.7 &40.5  &4.36 &55.4 &40.9 & 27.4\\

Context   & updated & updated & 19.7& 13.6 & 21.4 & 39.5 & 21.4 & 59.4 & 8.33 & 61.4  &43.6 & \textbf{32.1}\\
\hline

\end{tabular}
\end{center}
\caption{The mean relation accuracy on the PASCAL VOC 2012 dataset.}
\label{tbl:relation_acc_1}
\vspace{-5mm}
\end{table*}

\begin{table*}[t]
\scriptsize
\begin{center}
\begin{tabular}{c|c|c|c|c|c|c|c|c|c|c|c|c|c|c|c|c}

\hline
       &CNN & RsNN & behind & beside & fly & hold & play & in front & ride & sit & stand & under & walk & on & other & \textbf{mean} \\
\hline
Without  &partial fixed & updated &5.54& 9.24 &10.6 &27.3 &60.8 &5.93 &17.6 &39.7 &4.81 &17.4 &9.47&21.1&25.8&19.6\\

Context  &updated & updated &8.11& 11.3 & 16.1&37.0 &66.9 &7.20 &25.6 &41.4 &7.13 &23.0 &16.7&22.2&30.5&24.1\\

\hline

With     &partial fixed  & updated &7.33& 13.4 & 12.1&28.6 &61.6 &8.78 &23.0 &44.1 &4.41 &22.4 &9.69&23.4&24.9&21.8\\

Context  &updated & updated &10.1& 14.8 & 16.1&33.5&64.1 &12.8 &25.7 &49.7 &3.19 &20.8 &11.7&25.4&30.2&\textbf{24.5}\\
\hline

\end{tabular}
\end{center}
\caption{The mean relation accuracy on SYSU-Scenes.}
\label{tbl:relation_acc_2}
\vspace{-5mm}
\end{table*}

Table~\ref{tbl:result_end_to_end_learning} and Table~\ref{tbl:result_end_to_end_learning_2} report the results on the PASCAL VOC 2012 validation set and the SYSU-Scenes testing set. Table~\ref{tbl:relation_acc_1} and  Table~\ref{tbl:relation_acc_2} present the breakdown accuracy on relation categories. Fig.~\ref{fig:relation1} and Fig.~\ref{fig:relation2} show several examples of visualized scene parsing results on PASCAL VOC 2012 and SYSU-Scenes. The experiment results show that: (i) the incorporation of contextual information can benefit structure and relation prediction in terms of all the three performance metrics; (ii) joint optimization is very effective in improving structured scene parsing performance, no matter contextual information is considered or not.
Please refer to the supplementary materials for more successful and failure parsing results and our discussion on causes of failure.

\subsection{Inter-task Correlation}

Two groups of experiments are conducted to study the inter-task correlation of the two tasks: semantic labeling and scene structure generation (i.e., scene hierarchy construction and inter-object relation prediction). In the first group, we report the results with three different settings on the amount of strongly annotated data in semi-supervised learning of CNN-RsNN: i) zero strongly annotated image, ii) 280 strongly annotated images for PASCAL VOC 2012, and 500 strongly annotated images for SYSU-Scenes, and iii) 1464 strongly annotated images for PASCAL VOC 2012, and 1241 strongly annotated images for SYSU-Scenes. Other settings are the same with that described in Sec.~\ref{sub:semantic_Labelling}.

In the second group, we report the results with three different configurations on the employment of relation information in training CNN: i) zero relation, ii) relation category independent, and iii) relations category aware. In Configuration i), we ignore gradients from both the Scorer and the Categorizer sub-networks (see Sec.~\ref{sub:rnn_model}) of the RsNN model. In Configuration ii), we assume all relations are of the same class, and only back-propagate the gradients from the Scorer sub-network. In Configuration iii), we back-propagate the gradients from both the Scorer and the Categorizer sub-networks.

\begin{figure}[!htbp] \centering
\includegraphics[width=0.9\columnwidth]{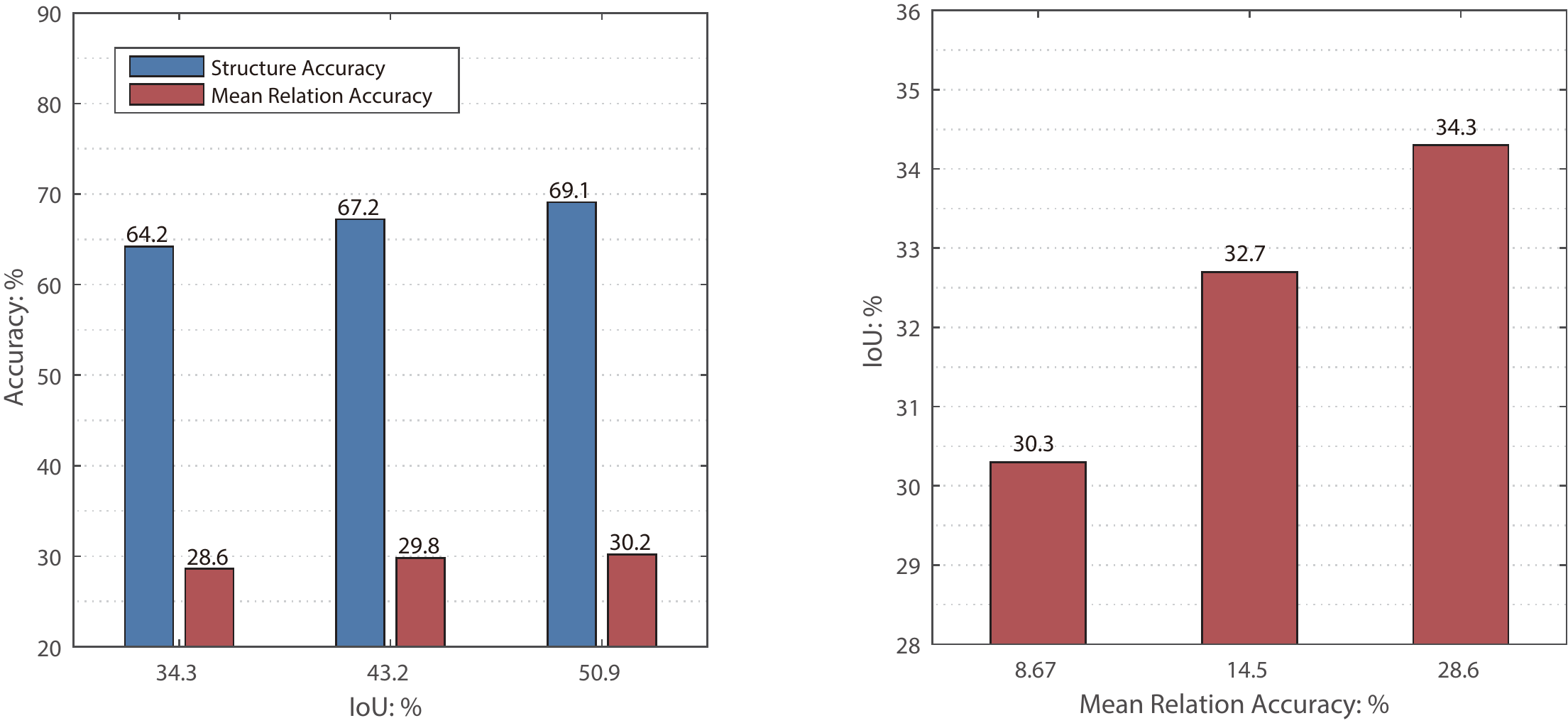}
\vspace{-3mm}
\caption{ Results of the inter-task correlation experiments on PASCAL 2012. The figure shows how segmentation and structure prediction task affect each other. Improving performance of one task results in improvement of the other. \textbf{The left} shows the effect of segmentation performance on relation and structure prediction based on the first group of experiments. \textbf{The right} shows the effect of relation prediction performance on semantic segmentation based on the second group of experiments. In practice, the segmentation performance is improved by adding more strongly annotated training data, while the performance of structure and relation prediction is improved by considering more types of relations.}
\label{fig:inter_task_correlation_voc}
\end{figure}

\begin{figure}[!htbp] \centering
\includegraphics[width=0.9\columnwidth]{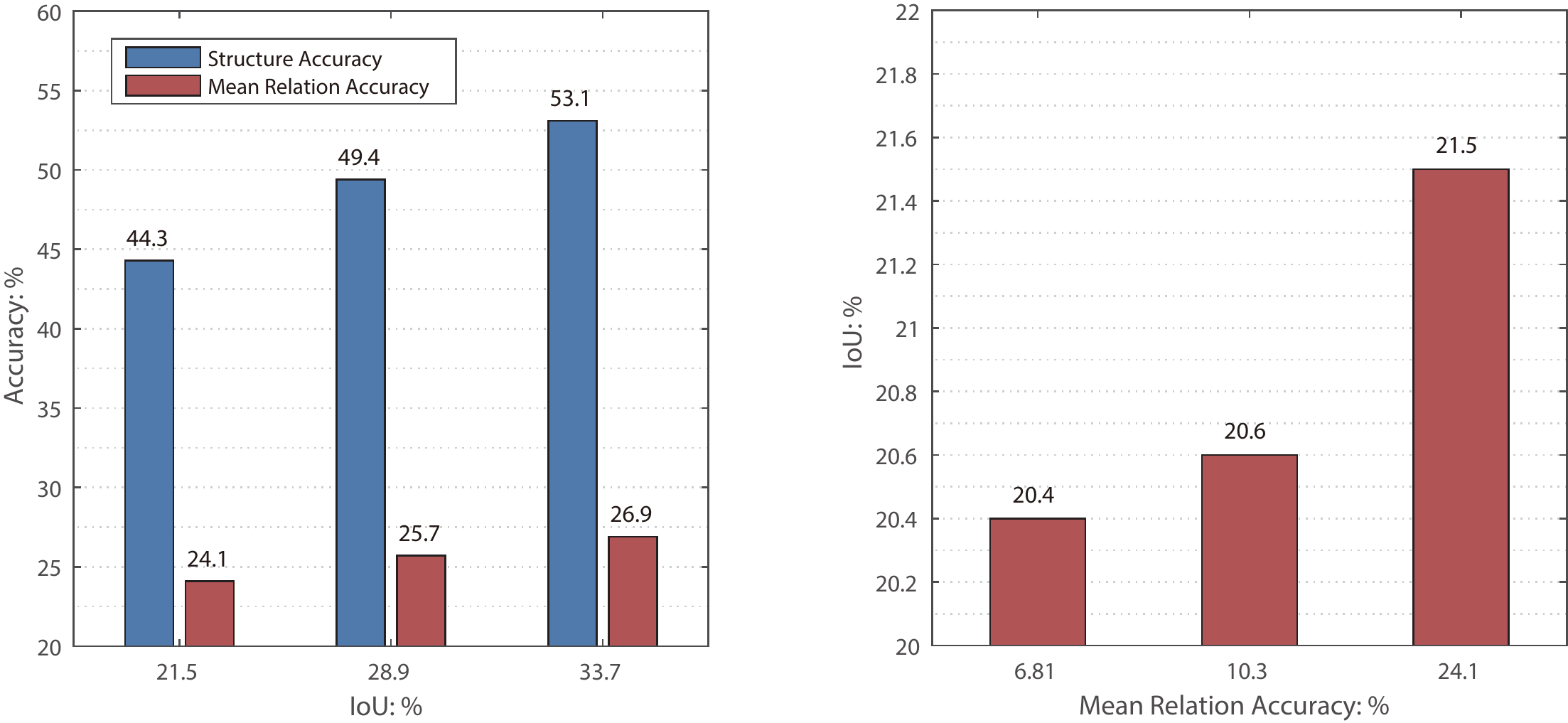}
\vspace{-3mm}
\caption{ Results of the inter-task correlation experiments on SYSU-Scenes. \textbf{The left} shows the effect of segmentation performance on relation and structure prediction experiments based on the first group of experiments. \textbf{The right} shows the effect of relation prediction performance on semantic segmentation based on the second group of experiments. }
\label{fig:inter_task_correlation_sysu}
\end{figure}

As shown in Fig.~\ref{fig:inter_task_correlation_voc} and Fig.~\ref{fig:inter_task_correlation_sysu}, the semantic labeling task is strongly correlated with the scene structure generation task. Increasing the amount of strongly annotated data and employing relation information can benefit both the semantic labeling and scene structure generation. As a result, the increase of relation/structure accuracy can result in a near-linear growth of semantic labeling accuracy.

We further study the correlation of two tasks under the full pixel supervision setting.
Different from the semi-supervised setting, we conduct the full pixel supervision without using extra data from SBD~\cite{sbd_dataset}.
Under this setting, we obtain two main observations as follows: (1) The introduction of full pixel supervision does benefit structure and relation prediction.
The accuracies of structure and relation prediction are $71.3\%$ and $39.5\%$ under the full pixel supervision, which are higher than the weakly-supervised setting with an obvious margin.
(2) Under the full pixel supervision, the further introduction of descriptive sentence contributes little in semantic labeling accuracy.
The mIoU of segmentation achieves $53.67\%$ on the PASCAL VOC \textit{val} dataset under the fully supervised setting, this value is improved only $0.13\%$ when image description is introduced to calculate the scene structure loss.
The results is natural since structure and relation prediction are performed after semantic labeling, and the pixel-wise classification loss is more effective than scene structure loss.

\section{Conclusion and Future Work}
\label{sec:conclusion}

In this paper, we have introduced a novel neural network model to address a fundamental problem of scene understanding, \textit{i.e.}, parsing an input scene image into a structured configuration including a semantic object hierarchy with object interaction relations. Our CNN-RsNN architecture integrates the convolutional neural networks and recursive neural networks for joint end-to-end training, and the two networks collaboratively handle the semantic object labeling and scene structure generation. To avoid expensively training our model with fully-supervised annotations, we have developed a weakly-supervised model training method by leveraging the sentence descriptions of training images. In particular, we distill rich knowledge from the sentence for discovering scene configurations. Experimental results have demonstrated the effectiveness of our framework by producing meaningful and structured scene configurations from scene images. We also release a new dataset to facilitate research on structured scene parsing, which includes elaborative annotations of scene configurations.

There are several directions in which we can do to extend this work. The first is to improve our framework by adding a component for recognizing object attributes in the scenes that corresponds the adjectives in the sentence descriptions. The second is to incorporate some instance segmentation~\cite{C:instnaceseg1,C:instnaceseg2,C:mask-rcnn} or object detection~\cite{J:FasterRCNN} model for instance level parsing. The third is to deeply combine our framework with state-of-the-art language processing techniques to improve the sentence parsing. Moreover, how to deal with the ambiguities of multiple sentence descriptions should be pursued.


\section*{Acknowledgement}

This work was supported by State Key Development Program under Grant 2016YFB1001004, the National Natural Science Foundation of China under Grant 61622214, and the Guangdong Natural Science Foundation Project for Research Teams under Grant 2017A030312006.

\ifCLASSOPTIONcaptionsoff
  \newpage
\fi

\section*{Supplementary Material}

\subsection{Dataset}
\label{sec:dataset}

\textbf{Sentence Annotation}. We asked 5 annotators to provide one descriptive sentence for each image in the PASCAL VOC 2012~\cite{pascal_voc} segmentation training and validation set.
Images from two sets are randomly partitioned into five subsets of equal size, each assigned to one annotator.
We provided annotators with a list of possible entity categories, which is the 20 defined categories in PASCAL VOC 2012 segmentation dataset.

\begin{figure*}[t]
\centering
\includegraphics[width=\linewidth]{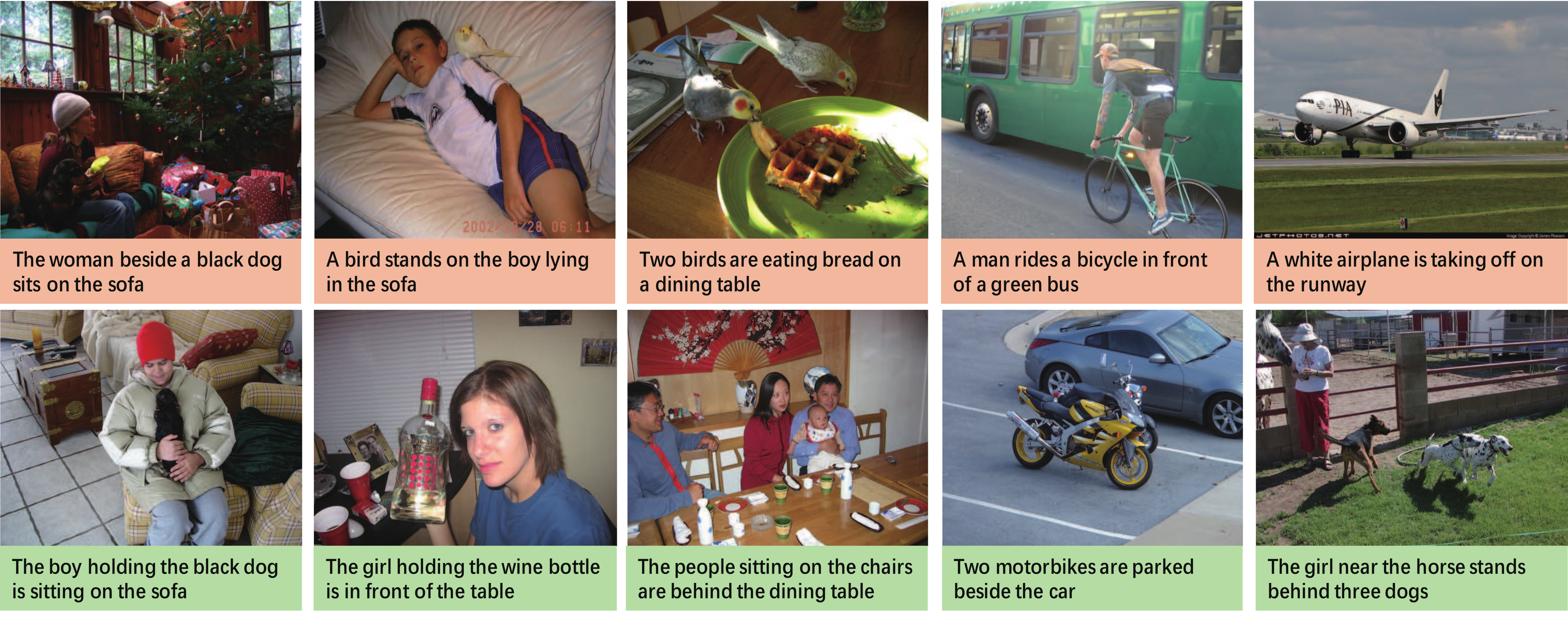}
\caption{Some pairs of images and annotated descriptions in PASCAL VOC 2012 dataset. Images in the first row are sampled from training set, while the second row's images are collected from the validation set.}
\label{fig:description}
\end{figure*}

We ask annotator to describe the main entities and their relations in the images.
We did not require them to describe all entities in images, as it would result in sentences being excessively long, complex and unnatural.
Fig.~\ref{fig:description} illustrates some pairs of images and annotated sentences in VOC 2012 \textit{train} and \textit{val} set.
%
%
For most images, both the objects and their interaction relations can be described with one sentence.
In particular, we summarize three significant annotation principles as follows:

\begin{itemize}

\item For the image with only an instance of some object category, e.g., the last image in the first row of Fig.~\ref{fig:description}, the sentence describes the relation between the object (i.e. airplane) and the background (i.e. runway);

\item For the instances from the same category with the same state, we describe them as a whole. Such as the forth image in the seconde row of Fig.~\ref{fig:description}, the annotation sentences is ``two motorbikes are parked beside the car".

\item For the instances from the same category with the different state, the annotator may only describe the most significant one. As to the third image in the second row of Fig.~\ref{fig:description}, the annotator describe the people sitting on the chairs but ignore the baby sitting on the adult.
\end{itemize}
We did not prohibit describing entities that did not belong to the defined categories, because they are necessary for natural expression.
But we will remove them in the process of generating semantic trees.

We annotate one sentence for each image because our method involves a language parser which produces one semantic tree for each sentence.
At this point, we are unable to generate one tree structure from multiple sentences.
Therefore, one sentence for each images is sufficient for our study.
To give more details of the image descriptions, we provide our sentence annotations of entire dataset in ``\textit{train\_sentences.txt}" and ``\textit{val\_sentences.txt}" as supplementary materials.

As described in the main paper, we parse sentences and convert them into semantic trees which consist of entities, scene structure and relations between entities.
Here we provide the list of 9 relation categories we defined: \textit{beside, lie, hold, ride, behind, sit on, in front of, on} and \textit{other}.
The label \textit{other} is assigned in the following two cases. (i) An entity has the relation with the background, which often happens at the last layer of the parsing structure. (ii) The \textit{other} relation is used as placeholder for the relation not identified as any of the 8 other relations

\begin{figure}[h]
\centering
\includegraphics[width=0.7\linewidth]{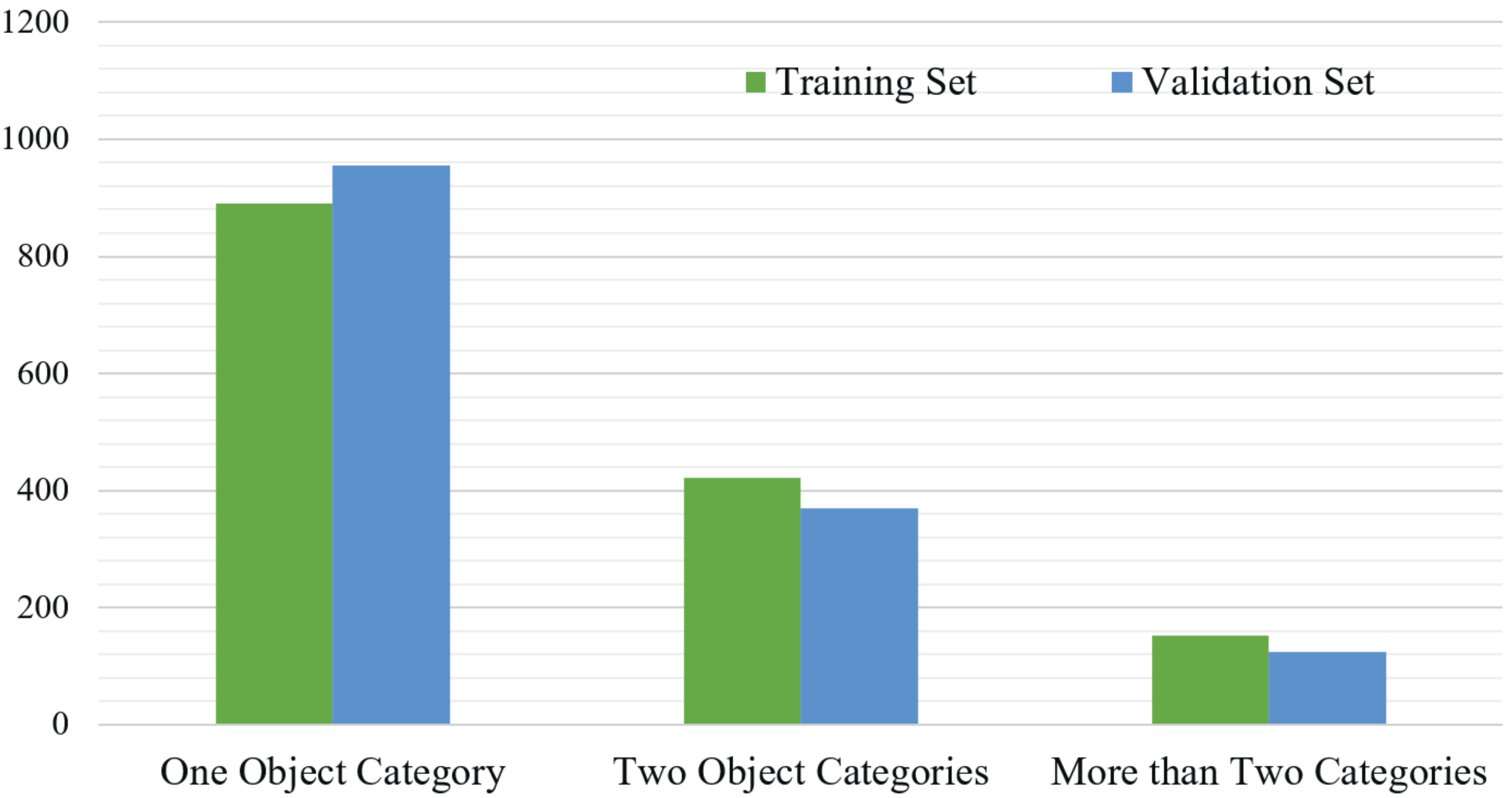}
    \caption{The number of object category of each image in VOC \textit{train} and \textit{val} dataset. The abscissa indicates the number of object categories in the image. The ordinate indicates the number of images. In each image, the number of interaction relations usually increases with the number of objects growing.}
\label{fig:ImageStas}
\end{figure}

\begin{figure}[t]
\centering
\includegraphics[width=0.7\linewidth]{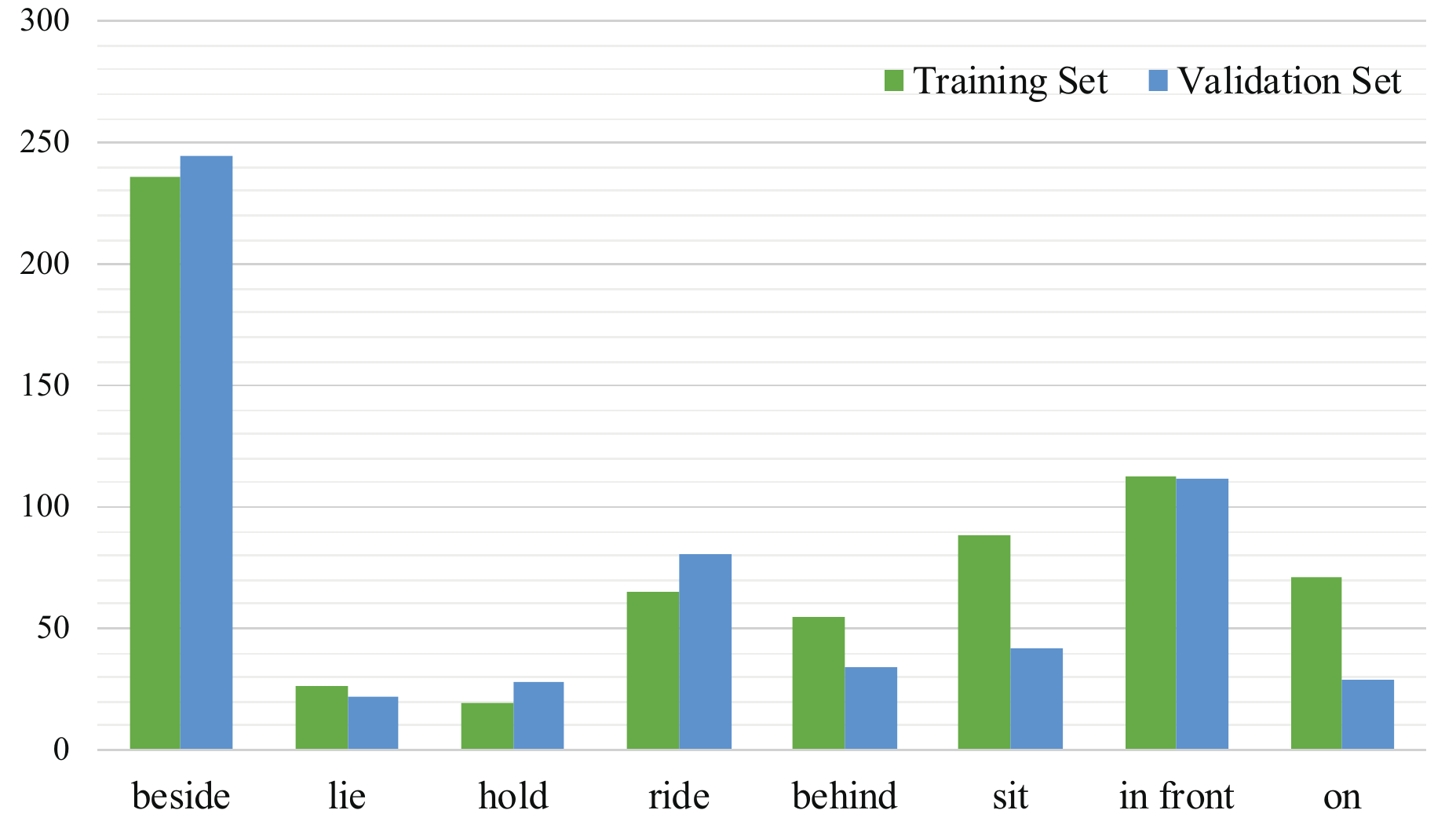}
    \caption{The number of occurrences of each relation category in the \textit{train} and \textit{val} dataset. Note that each image may contain multiply relations.  }
\label{fig:RelationStas}
\end{figure}

\textbf{Annotation Statistics}. Since the sentence annotations are not a standard part of the PASCAL VOC dataset, we give some statistical analysis of images and annotations in Fig.~\ref{fig:ImageStas} and Fig.~\ref{fig:RelationStas} to incorporate more information about our parsing task.
Fig.~\ref{fig:ImageStas} shows the number of object category of each image in VOC \textit{train} and \textit{val} dataset.
Obviously, for PASCAL VOC 2012 dataset, most images only contain one object category.
%
In order to construct the tree structure, we combine the foreground object and the background, and assign ``\textit{other}" as their relationship.
Another kind of images contain two or more object categories, and the number of relations in these images is greater than one.
As stated above, we combine the merged foreground objects and the background with the relation ``\textit{other}" at the last layer of the semantic tree.
According to the Fig.~\ref{fig:ImageStas}, the proportion of images with two or more object categories in the entire dataset is greater than $1/3$ (\textit{i.e.} $39.21\%$ for training set and $34.09\%$ for validation set).
Since the the number of interaction relations usually increases with the number of object growing, the total number of relations (except ``\textit{other}") in these images is more than $50\%$ of the entire dataset based on our sentence annotations.

Fig.~\ref{fig:RelationStas} reports the number of occurrences of each relation category in VOC \textit{train} and \textit{val} dataset.
The most common relation label is ``\textit{beside}", and the number of its occurrences is $236$ in training set and $245$ in validation set.
The label ``\textit{lie}" and ``\textit{hold}" are two least common labels, and occurrences times are around $20$ in both training and validation set.

\begin{figure*}[h]
\centering
\includegraphics[width=\linewidth]{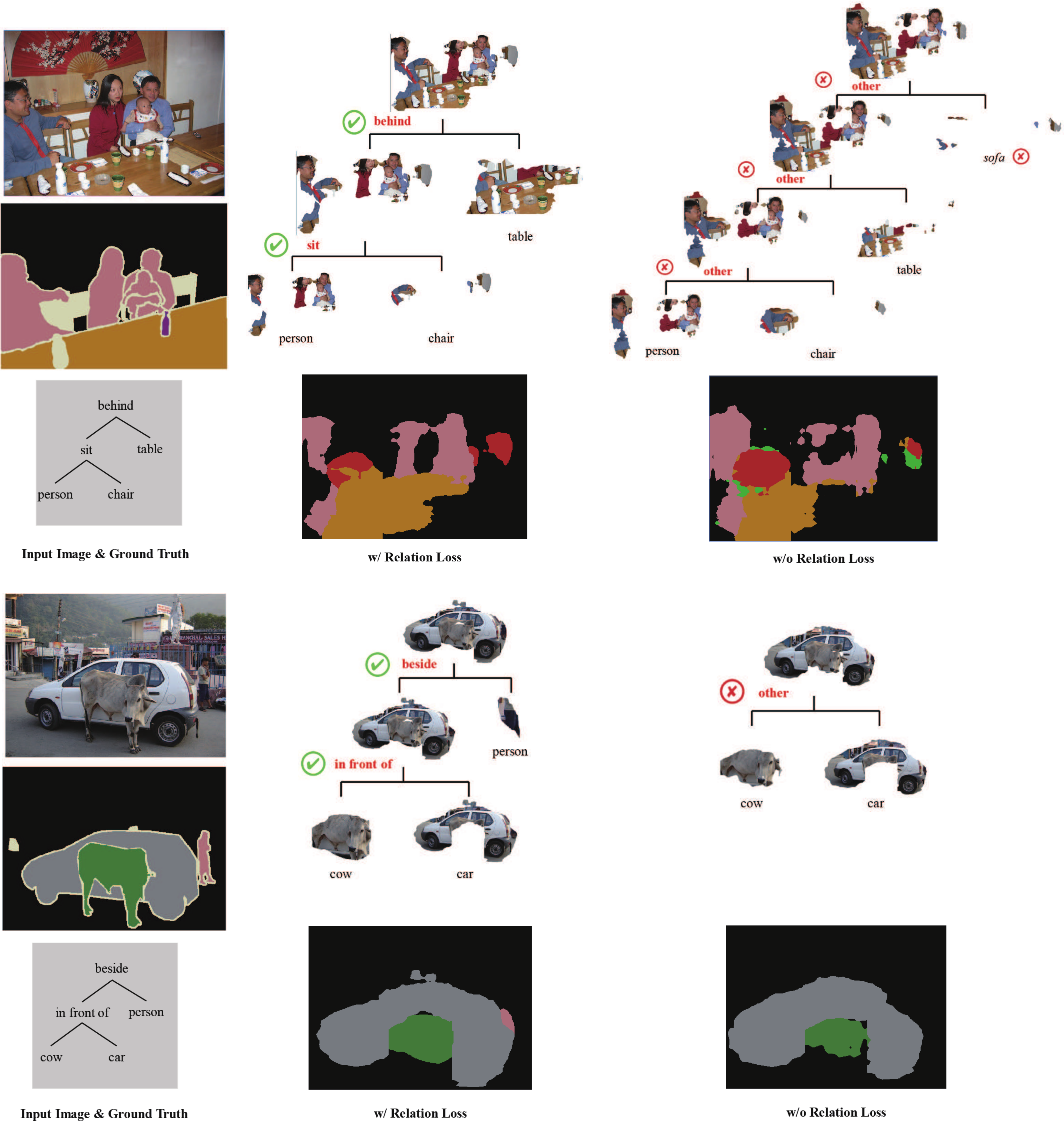}
    \caption{Some visualized semantic segmentation and scene structure prediction results with and without relation loss on PASCAL VOC 2012 val dataset. The first column shows the input images, the ground truth of semantic labeling and semantic trees. The second column gives the segmentation and structure prediction results with the relation loss (our method). In contract, the results without the relation loss are illustrated in the last column (like Socher et al. method).}
\label{fig:Comparison}
\end{figure*}


\subsection{Experiment Results}

\textbf{Analysis on Relation Loss}. We note that the RsNN model in previous works (e.g., Socher et al.~\cite{DBLP:Recursive_Socher}) only consider the structure supervision, but our model takes both structure and relation supervision during model training. To evaluate the performance of our method with and without relation supervision, we add some visualized results in Fig.~\ref{fig:Comparison}.
According to the figure, one can see that both of two methods learn the correct combination orders. However, our method can further predict the interaction relation between two merged object regions.
More importantly, the relation loss can also regularize the training process of CNN, which makes the segmentation model more effective to discover the small objects and eliminate the ambiguity.


\vspace{3mm}

\textbf{Analysis on Category Level Description}. Instead of instance-level parsing, this work aims to learn a CNN-RsNN
model for category-level scene parsing. When asking the annotator to describe the image,
some guidelines are introduced in Sec.\ref{sec:dataset}  to avoid instance-level descriptive sentences.
Under such circumstances, it is interesting to ask whether such annotation strategy are harmful to semantic labeling on images with multiple instances.
To answer this, we divide the VOC 2012 \emph{val} set into three subsets: (i) images with one instance from one object category, (ii) images with instances from multiple object categories, but only one instances from each category, and (iii) the others.
The mean IoU of our model on these three subsets are reported in Table~\ref{tbl:three_subsets}.
Although the number of object categories per image, the number of instances per category, and the number of images have the obvious difference among three subsets, the changes of mIoU remain in a small range.
It demonstrates that our category-level descriptions have little negative effect on semantic labeling results of images with multiple instances.

\begin{table}[t]
\begin{center}
\begin{tabular}{c|c|c|c}

\hline
Subset & (i) & (ii) & (ii) \\
\hline
Num. of Image &766 &266 &417 \\

mean IoU & 35.94\% & 33.19\% & 34.70\%  \\
\hline
\end{tabular}
\end{center}
\caption{Results on different subset of VOC 2012 \textit{val} under the weakly supervised learning.}
\label{tbl:three_subsets}
\end{table}

\vspace{3mm}

\begin{figure*}[t]
\centering
\includegraphics[width=\linewidth]{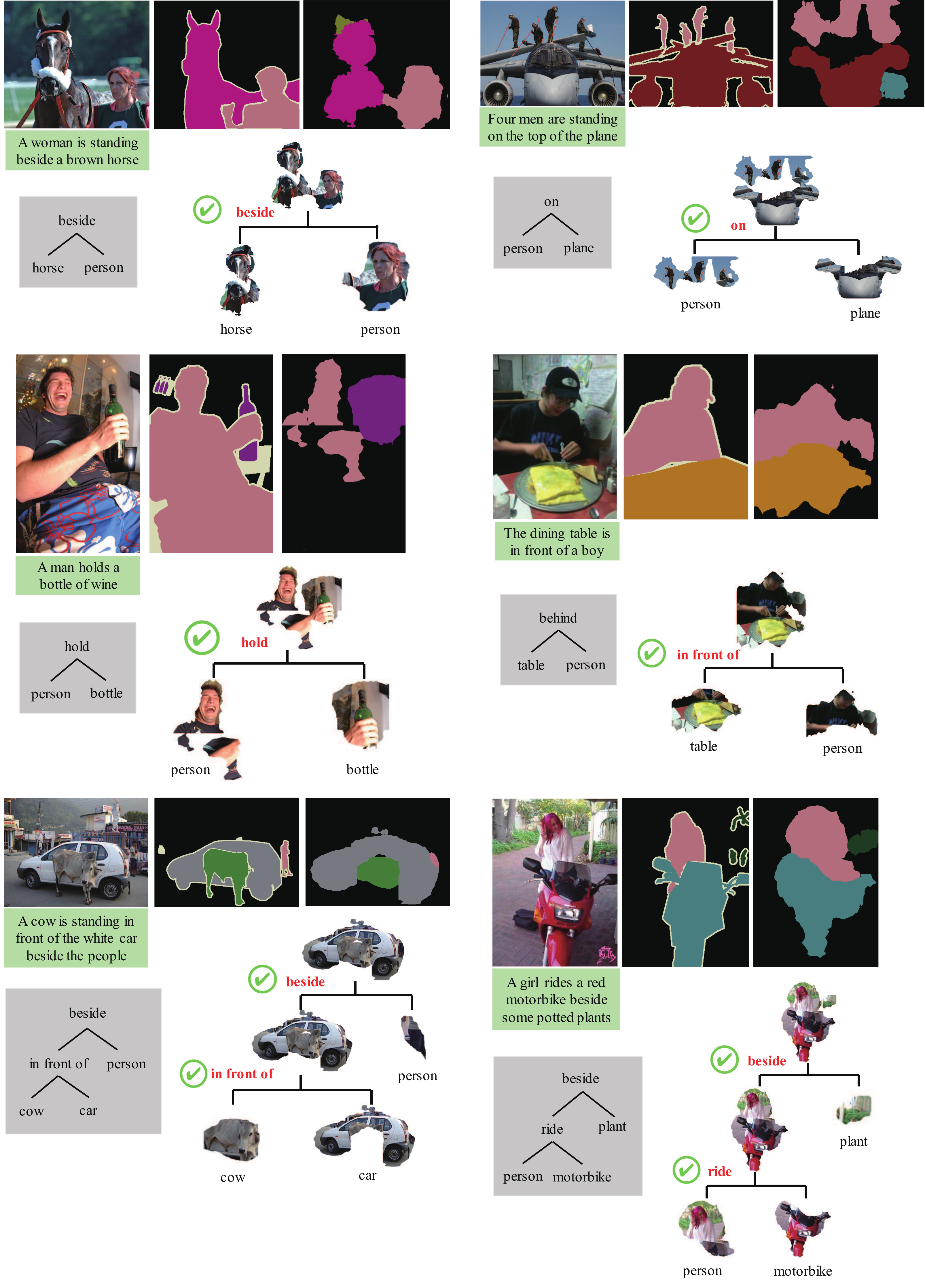}
    \caption{The visualized successful scene parsing results in PASCAL VOC 2012 dataset under the weakly supervised setting.}
\label{fig:good_result}
\end{figure*}

\begin{figure*}[h]
\centering
\includegraphics[width=\linewidth]{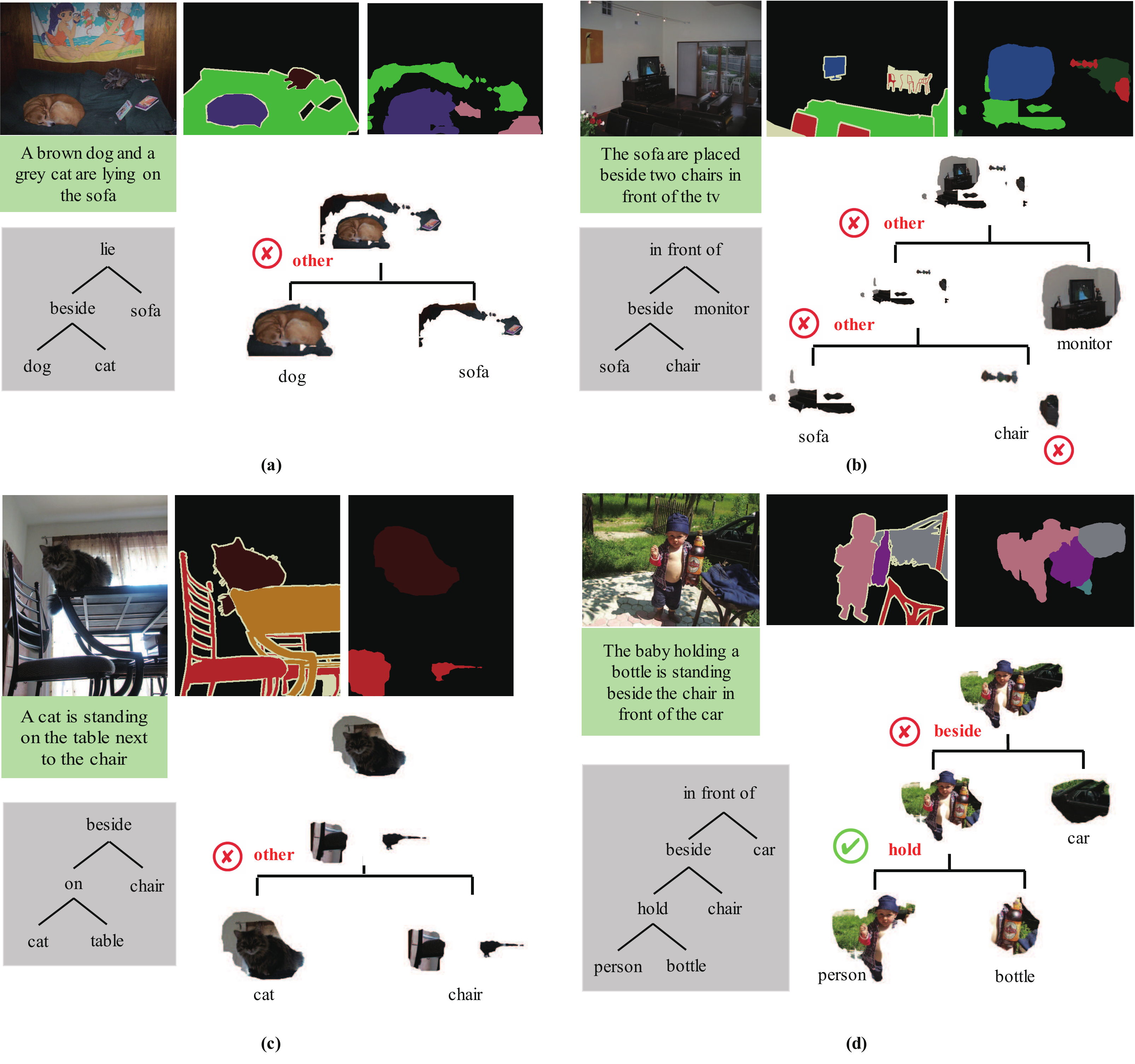}
    \caption{The visualized failure scene parsing results in PASCAL VOC 2012 dataset under the weakly supervised setting.}
\label{fig:failure_result}
\end{figure*}

\textbf{Analysis on Parsing Results}. To further investigate the performance of structure prediction, we provide some typical successful and failure cases of scene structure prediction in Fig.~\ref{fig:good_result} and Fig.~\ref{fig:failure_result}.
All of them are generated under the weakly supervised setting as described in the main paper.

We first show some \textit{successful} parsing results in Fig.~\ref{fig:good_result}. It is interesting to note that, our scene structure generation model is robust to small degree
of semantic labeling error.
As in the left image of the last row, even only a small part of the person is correctly labeled, both structure and relation prediction can be successfully predicted.
The relation categories in these examples cover most of the defined relations in this article.
Then, the \textit{failure} cases are illustrated in Fig.~\ref{fig:failure_result}.
According to this figure, the failure predictions usually happen in the following three cases.
(i) All of the structure and relation predictions are incorrect.
Fig.~\ref{fig:failure_result}-(a) and Fig.~\ref{fig:failure_result}-(c) illustrate such situation.
(ii) The structure is correct but the predicted relations are wrong.
Fig.~\ref{fig:failure_result}-(b) gives the example like this.
(iii) Both the structure and relation predictions are partially correct. Fig.~\ref{fig:failure_result}-(d) gives the example in such case.

According to the above discussion, one can see that the main cause of failure is the semantic labeling error, including seriously inaccurate labeling and complete failure in segmenting some object category.
Moreover, when the semantic labeling is inaccurate, the relation tends to be wrongly predicted as ¡±others¡± (see Fig.~\ref{fig:failure_result}-(a)(b)(c)).
When some object category is completely failed to be recognized, structure prediction is likely to be incorrect or partially incorrect (see Fig.~\ref{fig:failure_result}-(a)(d)).

%

\bibliographystyle{IEEEtran}
\bibliography{IEEEtran}

\vspace{-10mm}
\begin{IEEEbiography}[{\includegraphics[width=1in,height=1.25in,clip,keepaspectratio]{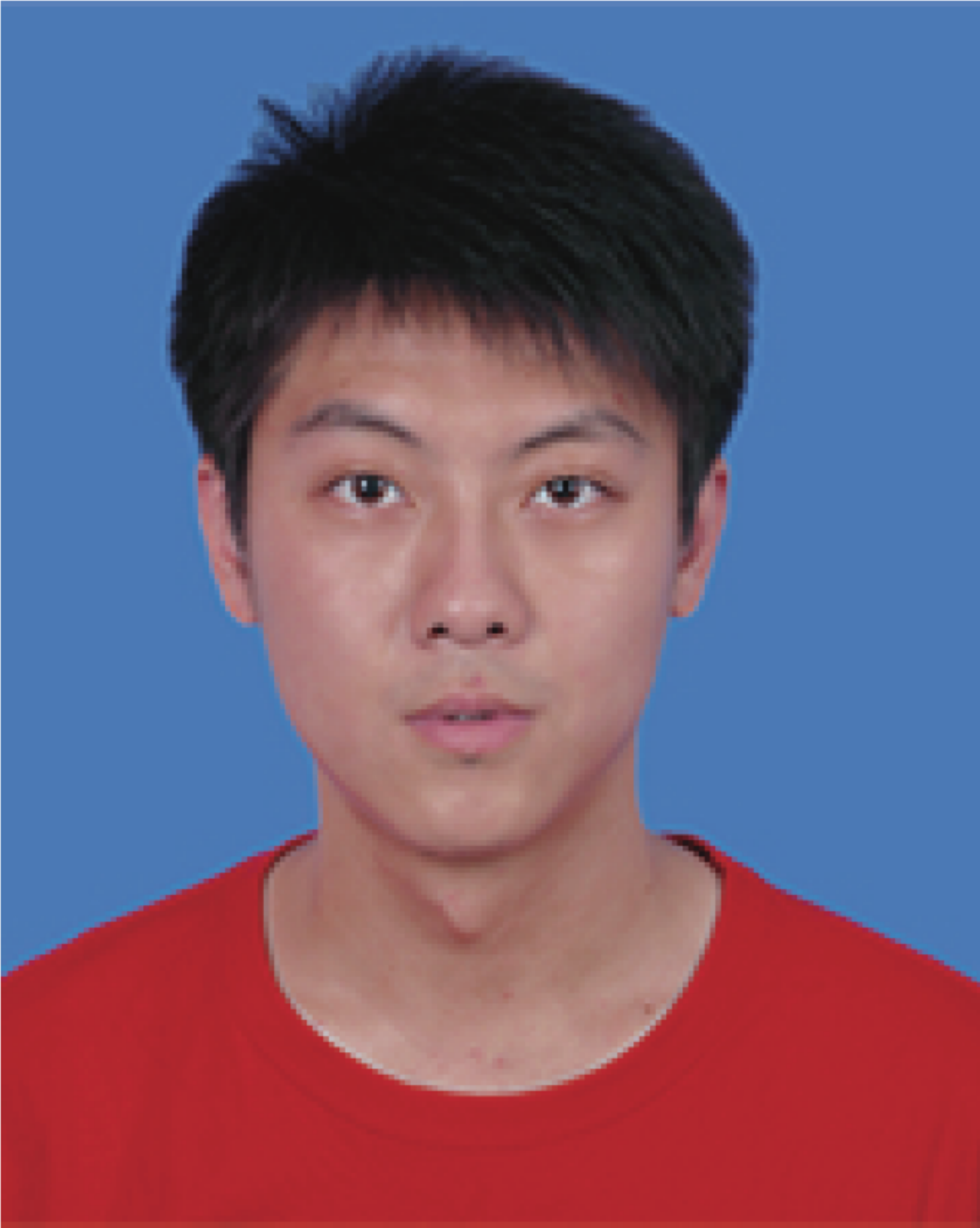}}]{Ruimao Zhang} is currently a postdoctoral fellow in the Department of Electronic Engineering, The Chinese University of Hong Kong (CUHK), Hong Kong, China. He received the B.E. and Ph.D. degrees from Sun Yat-sen University (SYSU), Guangzhou, China in 2011 and 2016, respectively. From 2013 to 2014, he was a visiting Ph.D. student with the Department of Computing, Hong Kong Polytechnic University (PolyU). His research interests include computer vision, deep learning and related multimedia applications. He currently serves as a reviewer of several academic journals, including IEEE Trans. on Neural Networks and Learning Systems, IEEE Trans. on Image Processing, IEEE Trans. on Circuits and Systems for Video Technology, Pattern Recognition and Neurocomputing.
\end{IEEEbiography}

\vspace{-10mm}
\begin{IEEEbiography}[{\includegraphics[width=1in,height=1.25in,clip,keepaspectratio]{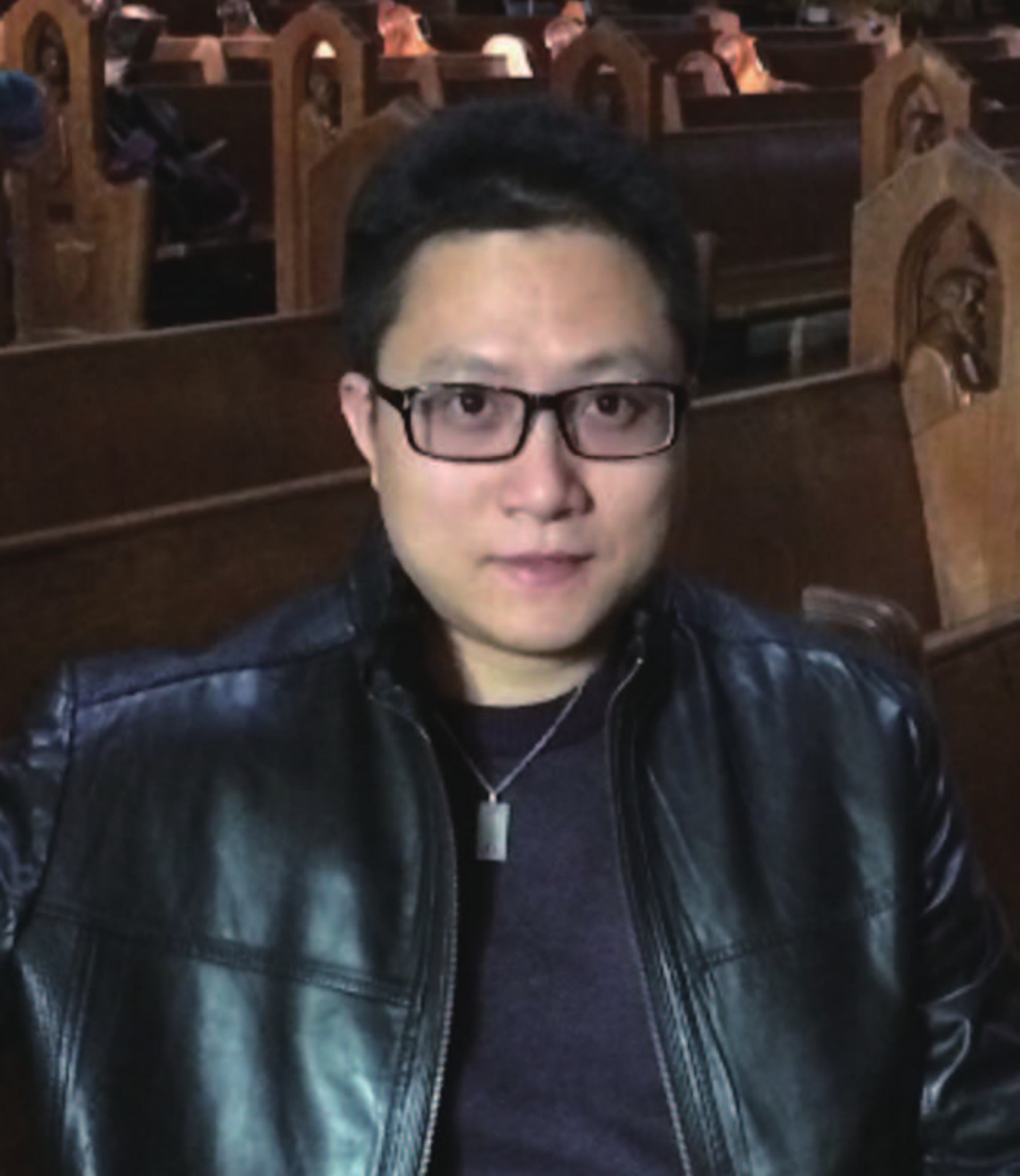}}]{Liang Lin}
(M'09, SM'15) is the Executive R$\&$D Director of SenseTime Group Limited and a full Professor of Sun Yat-sen University. He is the Excellent Young Scientist of the National Natural Science Foundation of China. From 2008 to 2010, he was a Post-Doctoral Fellow at University of California, Los Angeles. From 2014 to 2015, as a senior visiting scholar, he was with The Hong Kong Polytechnic University and The Chinese University of Hong Kong. He currently leads the SenseTime R$\&$D teams to develop cutting-edges and deliverable solutions on computer vision, data analysis and mining, and intelligent robotic systems. He has authorized and co-authorized on more than 100 papers in top-tier academic journals and conferences. He has been serving as an associate editor of IEEE Trans. Human-Machine Systems, The Visual Computer and Neurocomputing. He served as Area/Session Chairs for numerous conferences such as ICME, ACCV, ICMR. He was the recipient of Best Paper Runners-Up Award in ACM NPAR 2010, Google Faculty Award in 2012, Best Paper Diamond Award in IEEE ICME 2017, and Hong Kong Scholars Award in 2014. He is a Fellow of IET.
\end{IEEEbiography}

\vspace{-10mm}
\begin{IEEEbiography}[{\includegraphics[width=1in,height=1.25in,clip,keepaspectratio]{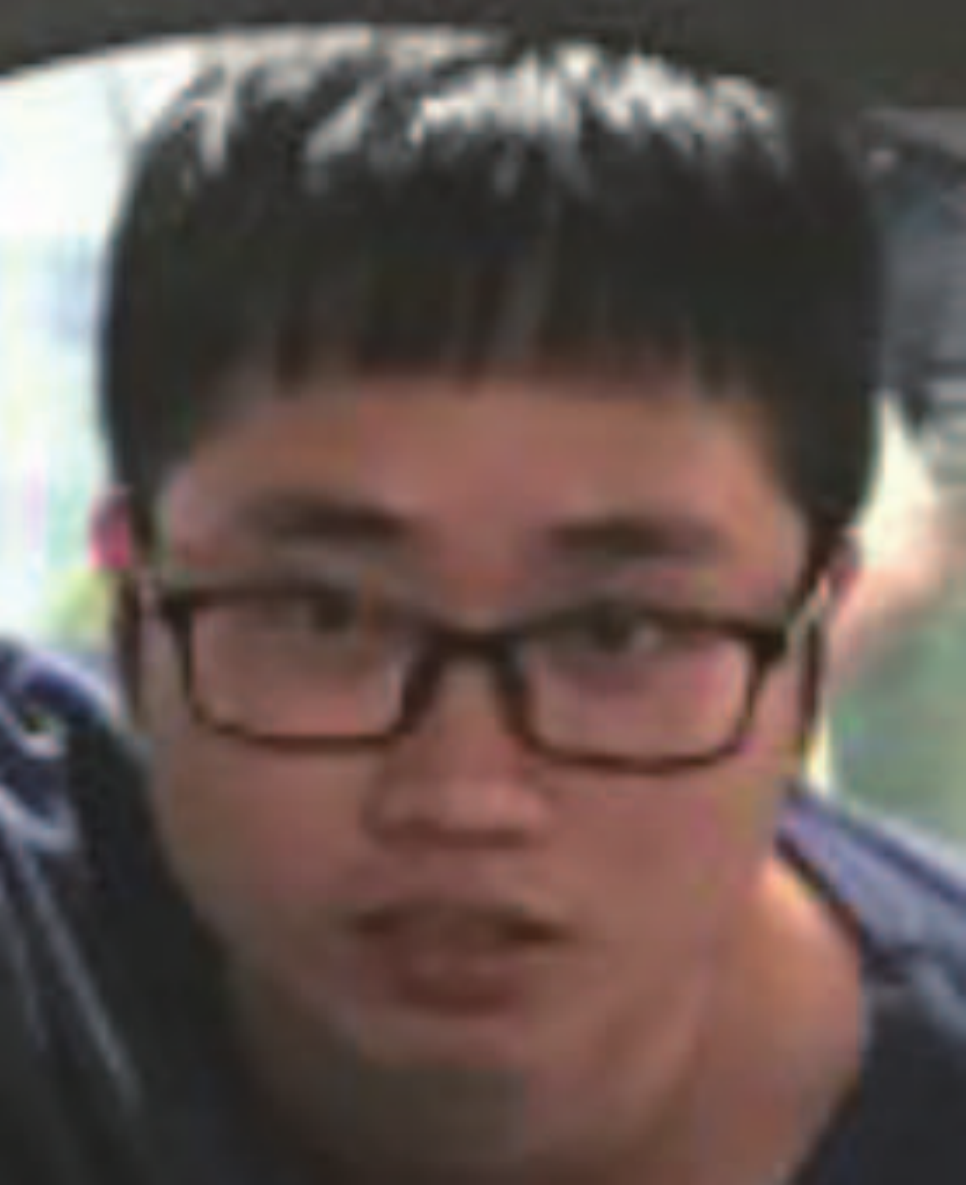}}]{Guangrun Wang}
is currently pursuing the Ph.D. degree in the School of Data and Computer Science, Sun Yat-sen University. He received the B.E. degree from Sun Yat-sen University, Guangzhou, China, in 2014. From Jan 2016 to Aug 2017, he was a visiting scholar with the Department of Information Engineering, the Chinese University of Hong Kong.
His research interests include computer vision and machine learning.
\end{IEEEbiography}

\vspace{-10mm}
\begin{IEEEbiography}[{\includegraphics[width=1in,height=1.25in,clip,keepaspectratio]{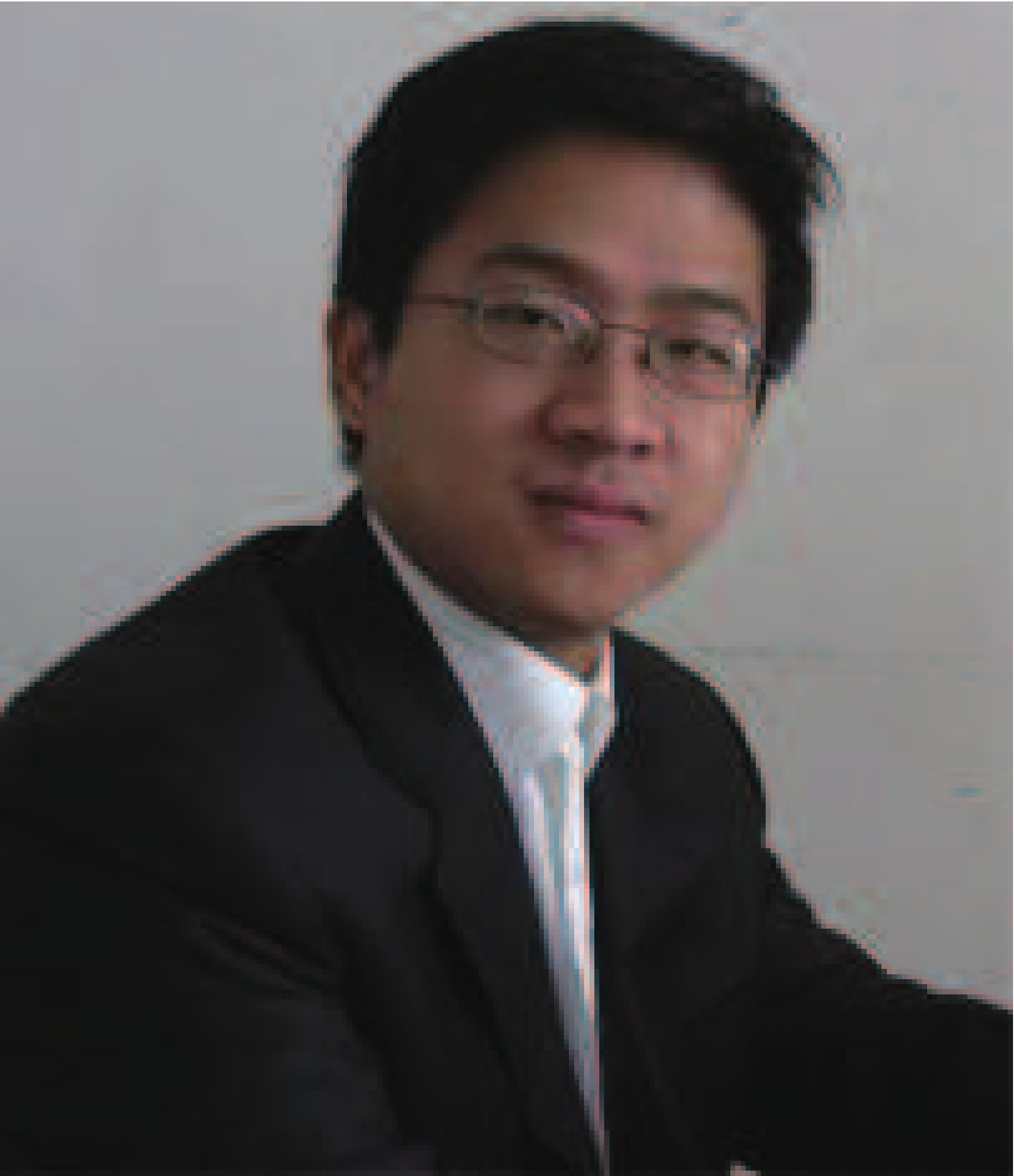}}]{Meng Wang}
is a professor at the Hefei University of Technology, China. He received his B.E. degree and Ph.D. degree in the Special
Class for the Gifted Young and the Department of Electronic Engineering and Information Science from the University of Science and
Technology of China (USTC), Hefei, China, in 2003 and 2008, respectively. His current research interests include multimedia
content analysis, computer vision, and pattern recognition. He has authored more than 200 book chapters, journal and conference papers in
these areas. He is the recipient of the ACM SIGMM Rising Star Award 2014. He is an associate editor of IEEE Transactions on Knowledge and Data
Engineering (IEEE TKDE), IEEE Transactions on Circuits and Systems for Video Technology (IEEE TCSVT), and IEEE Transactions on Neural Networks and Learning Systems (IEEE TNNLS).
\end{IEEEbiography}

\vspace{-10mm}
\begin{IEEEbiography}[{\includegraphics[width=1in,height=1.25in,clip,keepaspectratio]{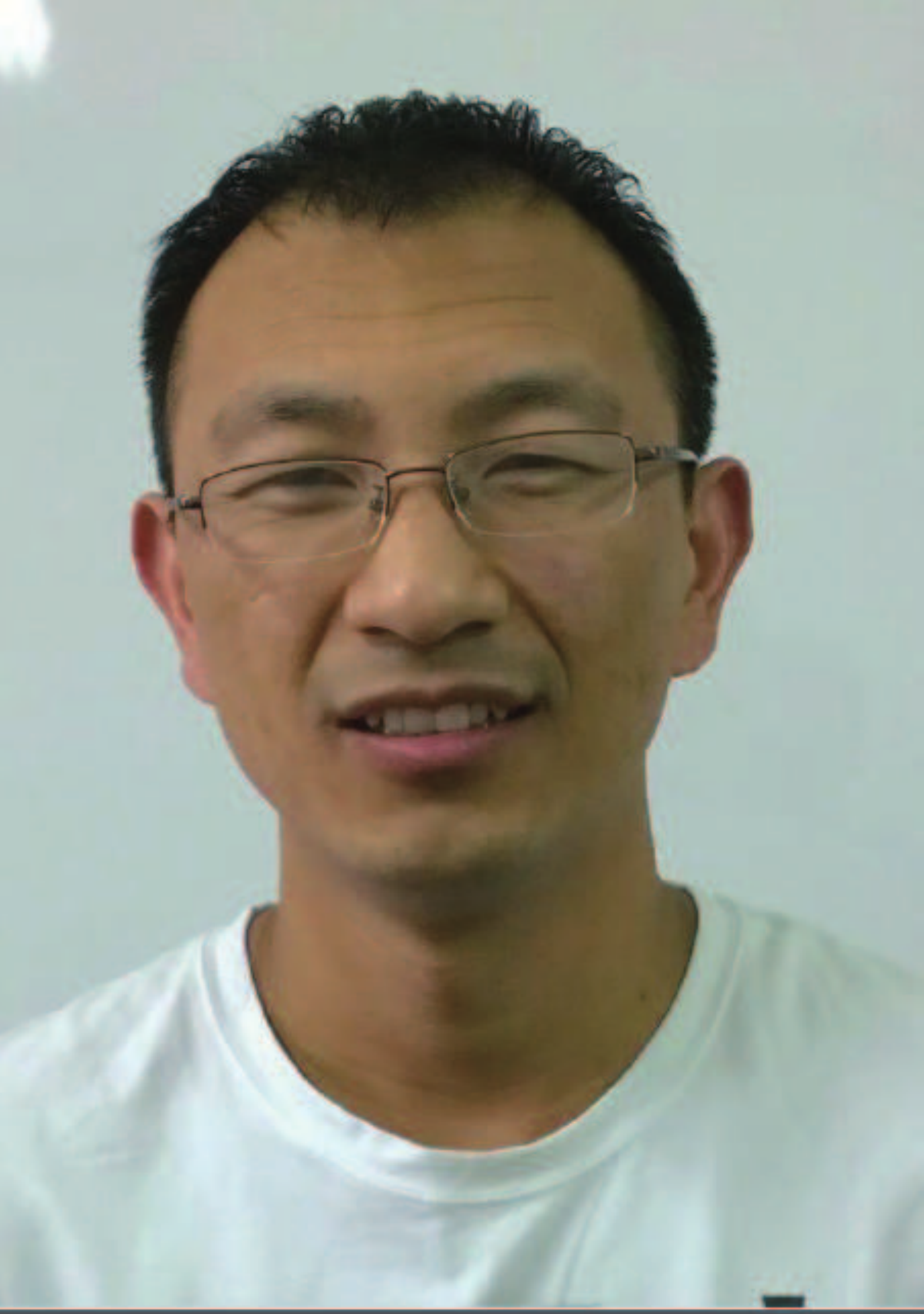}}]{Wangmeng Zuo}
received the Ph.D. degree in computer application technology from the Harbin Institute of Technology, Harbin, China, in 2007. He is currently a Professor in the School of Computer Science and Technology, Harbin Institute of Technology. His current research interests include image enhancement and restoration, object detection, visual tracking, and image classification.He has published over 60 papers in top-tier academic journals and conferences. 

\end{IEEEbiography}

\end{document}